\RequirePackage{expl3}
\documentclass{template/ieeeaccess}
\usepackage[dvipdfmx]{graphicx} 
\usepackage{bmpsize}

\usepackage{caption}
\usepackage{amsthm}
\usepackage{algorithm, algorithmicx, algpseudocode}
\usepackage{mathrsfs}
\usepackage{times}
\usepackage{multicol}
\usepackage[caption=false, font=footnotesize]{subfig}
\usepackage[utf8]{inputenc}
\usepackage[T1]{fontenc}
\usepackage{textcomp}
\usepackage{balance}
\usepackage{multirow}
\usepackage{booktabs}
\usepackage{dblfloatfix} 
\usepackage[normalem]{ulem} % for strikethrough font \sout{Hellow world}
\usepackage{bbold}
\usepackage{makecell}
\usepackage{hhline} % double-line in a table

\usepackage{comment}
\usepackage{amsmath,amssymb,enumerate, amsfonts}

\usepackage{setspace}

\usepackage{xparse}
\usepackage{lipsum}
\usepackage{soul} 
\usepackage{times}
\usepackage{cite} % to group references
\usepackage{fixltx2e}
\usepackage[usenames,dvipsnames,svgnames,table, xcdraw]{xcolor}
\usepackage{balance}

\newtheorem{remark}{Remark}

\graphicspath{
    {graphics/}
}
\def\realnumbers{\mathbb{R}}
\def\real{\mathbb{R}}
\DeclareMathOperator*{\argmin}{arg\,min}
\DeclareMathOperator*{\argmax}{arg\,max}
\definecolor{note}{rgb}{0.3,0.7,0.25}
\definecolor{rephase}{rgb}{0.15,0.7,0.15}
\definecolor{bag}{rgb}{0.6,0.6,0.2}

\newcommand{\lidar}{LiDAR~}
\newcommand{\lidars}{LiDARs~}

\newcommand{\pc}{\mathcal{PC}}
\newcommand{\vx}{\mathcal{V}}

\newcommand{\Velodyne}{32-Beam Velodyne ULTRA Puck LiDAR~}

\newcommand{\transpose}{\mathsf{T}}

\newcommand{\norm}[1]{\left\lVert#1\right\rVert}

\DeclareDocumentCommand{\vector}{ O{} }{\mathrm{vec}(#1)}
\DeclareDocumentCommand{\zeros}{ O{} }{\textbf{0}_{#1}}
\DeclareDocumentCommand{\X}{ O{} O{} }{\textbf{X}_{#1}^{#2}}
\DeclareDocumentCommand{\A}{ O{} O{} }{\textbf{A}_{#1}^{#2}}
\DeclareDocumentCommand{\H}{ O{} O{} }{\textbf{H}_{#1}^{#2}}
\DeclareDocumentCommand{\pre}{ O{} O{} }{{}_{#1}^{#2}}
\DeclareDocumentCommand{\P}{ O{} O{} }{\textbf{P}_{#1}^{#2}}
\DeclareDocumentCommand{\R}{ O{} O{} }{\textbf{R}_{#1}^{#2}}
\DeclareDocumentCommand{\T}{ O{} O{} }{\textbf{T}_{#1}^{#2}}
\DeclareDocumentCommand{\V}{ O{} O{} }{\textbf{V}_{#1}^{#2}}
\DeclareDocumentCommand{\Y}{ O{} O{} }{\textbf{Y}_{#1}^{#2}}
\DeclareDocumentCommand{\U}{ O{} O{} }{\textbf{U}_{#1}^{#2}}

\begin{document}
\history{Date of publication xxxx 00, 0000, date of current version xxxx 00, 0000.}
\doi{10.1109/ACCESS.2017.DOI}

\title{Improvements to Target-Based 3D LiDAR to Camera Calibration}
\author{\uppercase{Jiunn-Kai Huang}\authorrefmark{1} and \uppercase{Jessy W. Grizzle}\authorrefmark{1}
\address[1]{Robotics Institute, University of Michigan, Ann Arbor, MI 48109, USA.}}
\corresp{Corresponding author: Jiunn-Kai Huang (e-mail: bjhuang@ umich.edu).}

\tfootnote{Funding for this work was provided by the Toyota Research Institute (TRI) under award number N021515. Funding for J. Grizzle was in part provided by TRI and in part by NSF Award No.~1808051. This article solely reflects the opinions and conclusions of its authors and not the funding entities. Dr. M Ghaffari offered advice during the course of this project. The first author thanks Wonhui Kim for useful conversations and Ray Zhang for generating the semantic image.}

\markboth
{JK Huang \headeretal: Improvements to Target-Based 3D LiDAR to Camera Calibration}
{JK Huang \headeretal: Improvements to Target-Based 3D LiDAR to Camera Calibration}

\begin{abstract}
The rigid-body transformation between a LiDAR and monocular camera is required for sensor fusion tasks, such as SLAM. While determining such a transformation is not considered glamorous in any sense of the word, it is nonetheless crucial for many modern autonomous systems. Indeed, an error of a few degrees in rotation or a few percent in translation can lead to 20 cm reprojection errors at a distance of 5 m when overlaying a LiDAR image on a camera image. The biggest impediments to determining the transformation accurately are the relative sparsity of LiDAR point clouds and systematic errors in their distance measurements. This paper proposes (1) the use of targets of known dimension and geometry to ameliorate target pose estimation in face of the quantization and systematic errors inherent in a LiDAR image of a target, (2) a fitting method for the LiDAR to monocular camera transformation that avoids the tedious task of target edge extraction from the point cloud, and (3) a ``cross-validation study'' based on projection of the 3D LiDAR target vertices to the corresponding corners in the camera image. The end result is a 50\% reduction in projection error and a 70\% reduction in its variance with respect to baseline.
\end{abstract}


% \titlepgskip=-15pt

\maketitle

%%%%%%%%%%%%%%%%%%%%%%%%%%
\section{Introduction and Related Work}
\label{sec:intro}
\begin{figure}[t]
\centering
\includegraphics[width=1\columnwidth, bb=0 0 1595 618]{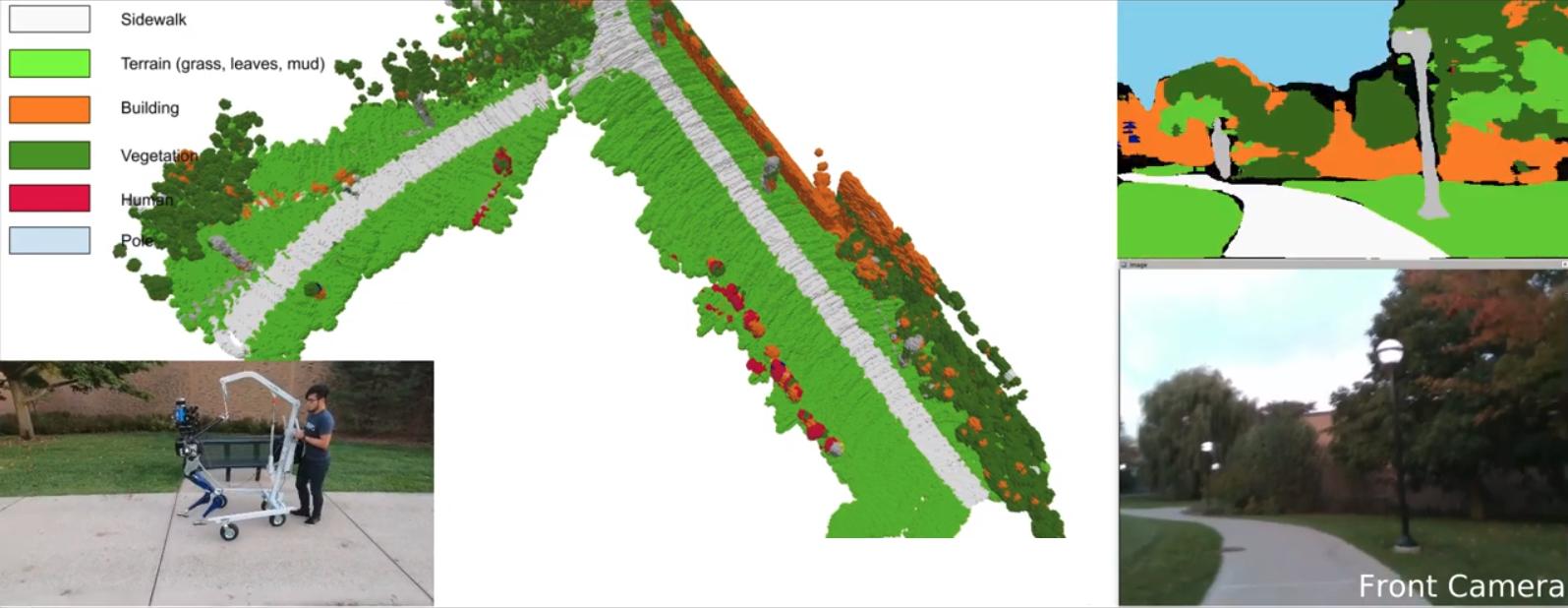}
\caption{Building a semantic map for autonomous navigation. Once the transformation from LiDAR to camera is known, it is possible to fuse the LiDAR and RGB-camera information to build a 3D-map that is annotated with semantic information. The lower right shows a single camera image; its segmentation with semantic labels is shown in the upper right. Synchronized LiDAR points are mapped onto the camera image, associated to a semantic label, and then re-projected to a local frame \cite{hartley2019contact} along with the semantic labels to create a 3D semantically-labeled  map \cite{Lu2020BKI}, shown in the upper left. The sidewalk is marked in white, the grass in yellow-green, and trees in dark green. In the lower left, Cassie is using the map to plan a path around the North Campus Wave Field, while staying on the sidewalk.}
\label{fig:semantic}
\end{figure}

The desire to produce 3D-semantic maps \cite{Lu2020BKI} with our Cassie-series bipedal robot~\cite{gong2019feedback} has motivated us to fuse 3D-LiDAR and RGB-D
monocular camera data for autonomous navigation
\cite{CassieAutonomy2019ExtendedEdition}. Indeed, by mapping spatial LiDAR points
onto a segmented and labeled camera image, one can associate the label of a pixel (or
a region about it) to the LiDAR point as shown in Fig.~\ref{fig:semantic}. An error of a few degrees in rotation or a few percent in translation in the estimated rigid-body transformation between LiDAR and camera can lead to 20 cm reprojection errors at a distance of 5 m when overlaying a LiDAR point cloud on a camera image. Such errors will lead to navigation errors. 

In this paper, we assume that the intrinsic calibration of the two sensors has
already been done~\cite{mirzaei20123d} and focus on obtaining the rigid-body
transformation, i.e., rotation matrix and translation, between a LiDAR and camera.
This is a well studied problem with a rich literature that can be roughly divided
into methods that do not require targets:
\cite{pandey2012automatic,taylor2013automatic,jeong2019road, jiang2018line, erke2020fast, 
zhen2019joint} and those that do: \cite{gong20133d,dhall2017lidar,verma2019automatic,jiao2019novel,kim2019extrinsic, guindel2017automatic, garcia2013lidar, mishra2020extrinsic, xue2019automatic, vaida2019automatic, an2020geometric}. While many of the existing target-based methods may work well, in practice, they require many manual steps, such as carefully measuring different parts of the targets, parsing edge-points, and edge-points pairing inspection. Our method avoids these manual steps and is applicable to planar polygonal targets, such as checkerboards, triangles, and diamonds.

In target-based methods, one seeks to estimate a set of target features (e.g., edge
lines, normal vectors, vertices) in the LiDAR's point cloud and the camera's image
plane. If ``enough'' independent correspondences can be made, the LiDAR to camera
transformation can be found by Perspective-n-Point (PnP) as in
\cite{lepetit2009epnp}, that is, through an optimization problem of the form
\begin{equation}
\label{eq:ConceptualProblem} 
(R_{C}^{L*}, T_{C}^{L*})= \mathrm{arg} \min_{(R,T)}{\sum_i \mathrm{dist}(P(H_{L}^{C}(X_i)),Y_i)}, \end{equation} 
where $X_i$ are
the (homogeneous) coordinates of the LiDAR features, $Y_i$ are the coordinates of the
camera features, $P$ is the often-called ``projection map'', $H_{L}^{C}$ is the
(homogeneous representation of) the LiDAR-frame to camera-frame transformation with
rotation matrix $R_{C}^{L}$ and translation $T_{C}^{L}$, and $\mathrm{dist}$ is a
distance or error measure.

\subsection{Rough Overview of the Most Common Target-based Approaches}
The works closest to ours are \cite{liao2018extrinsic,zhou2018automatic}. Each of
these works has noted that rotating a square target so that it presents itself as a
diamond can help to remove pose ambiguity due to the spacing of the ring lines; in
particular, see Fig.~2 in \cite{liao2018extrinsic} and Fig.~1 in
\cite{zhou2018automatic}. More generally, we recommend the literature overview in
\cite{liao2018extrinsic} for a recent, succinct survey of LiDAR to camera
calibration.  

 %trim={<left> <lower> <right> <upper>}

The two most common sets of features in the area of target-based calibration are (a)
the 3D-coordinates of the vertices of a rectangular or triangular planar target, and
(b) the normal vector to the plane of the target and the lines connecting the
vertices in the plane of the target. Mathematically, these two sets of data are
equivalent: knowing one of them allows the determination of the other. In practice,
focusing on (b) leads to use of the SVD to find the normal vector and more broadly to
least squares line fitting problems \cite{zhou2018automatic}, while (a) opens up
other perspectives, as highlighted in \cite{liao2018extrinsic}.

% \begin{figure}[t]%
% \centering
% \subfloat{%
% \label{fig:semanticPC}%
% \centering
% \includegraphics[width=0.50\columnwidth, trim={14.1cm 2cm 1cm 8.2cm},clip]{semanticPC3.png}}%
% \hspace{5pt}%
% \subfloat{%
% \label{fig:semanticImg}%
% \centering
% \includegraphics[width=0.45\columnwidth, trim={0.00cm 0cm 0cm 0.cm},clip]{semanticImg.png}}%
% \caption[]{\jwg{Left image needs to be more fun and interesting} Using the obtained transformation, LiDAR points are mapped onto a semantically segmented image. Each point is associated with the label of a pixel. The road is marked as white; static objects such buildings as orange; the grass as yellow-green, and dark green indicates trees. \jwg{Add citation to YouTube video with nice map}}%
% \label{fig:semantic}%
% \squeezeup
% \end{figure}

 %trim={<left> <lower> <right> <upper>}
\begin{figure}[t]%
\centering
\subfloat[]{%
\label{fig:FigureLiDARScanExample}%
\includegraphics[width=0.48\columnwidth, trim={10cm 2.5cm 12cm 1cm},clip]{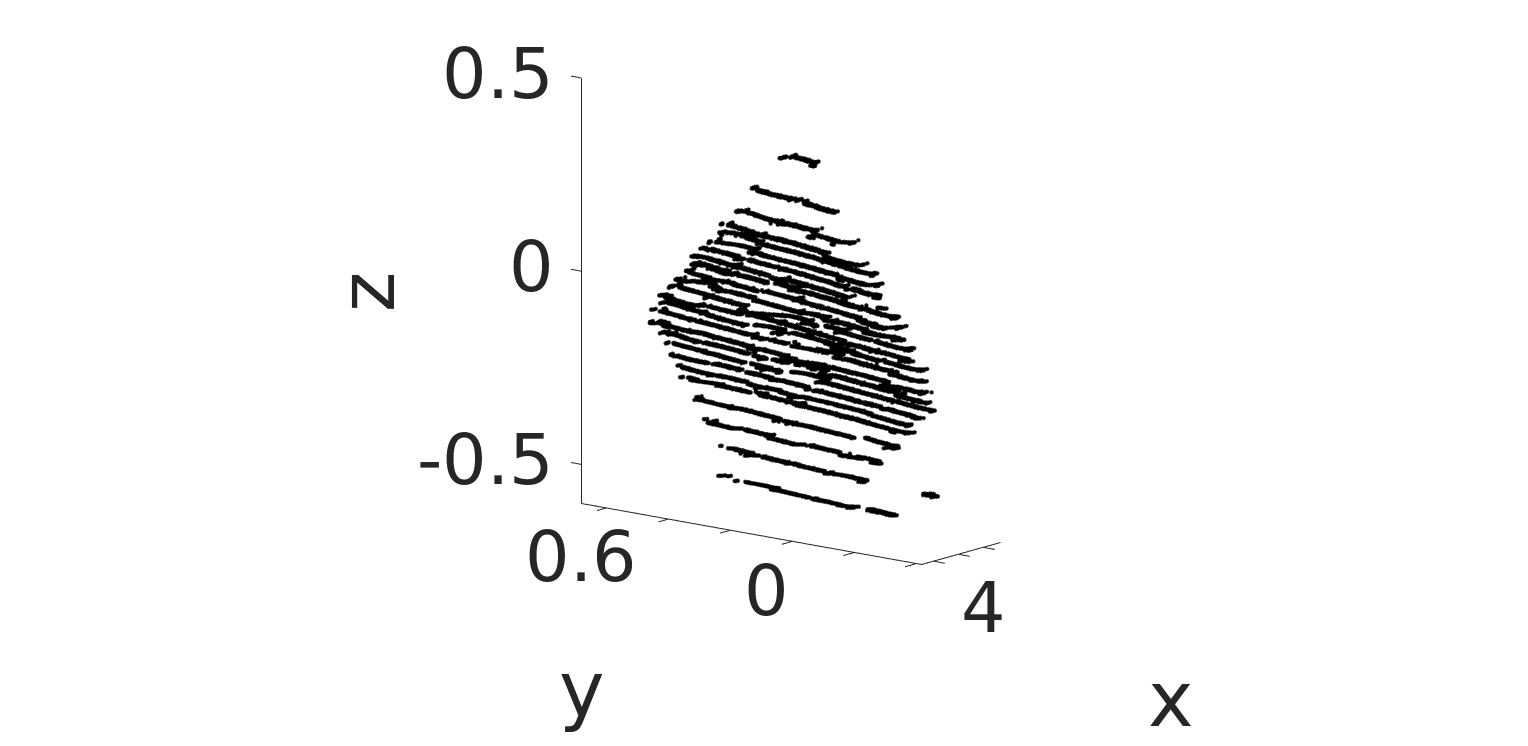}}%
\hspace{8pt}%
\subfloat[]{%
\label{fig:FigureLiDARScanExample3Rings}%
\includegraphics[width=0.47\columnwidth, trim={10cm 0cm 11cm 2.0cm},clip]{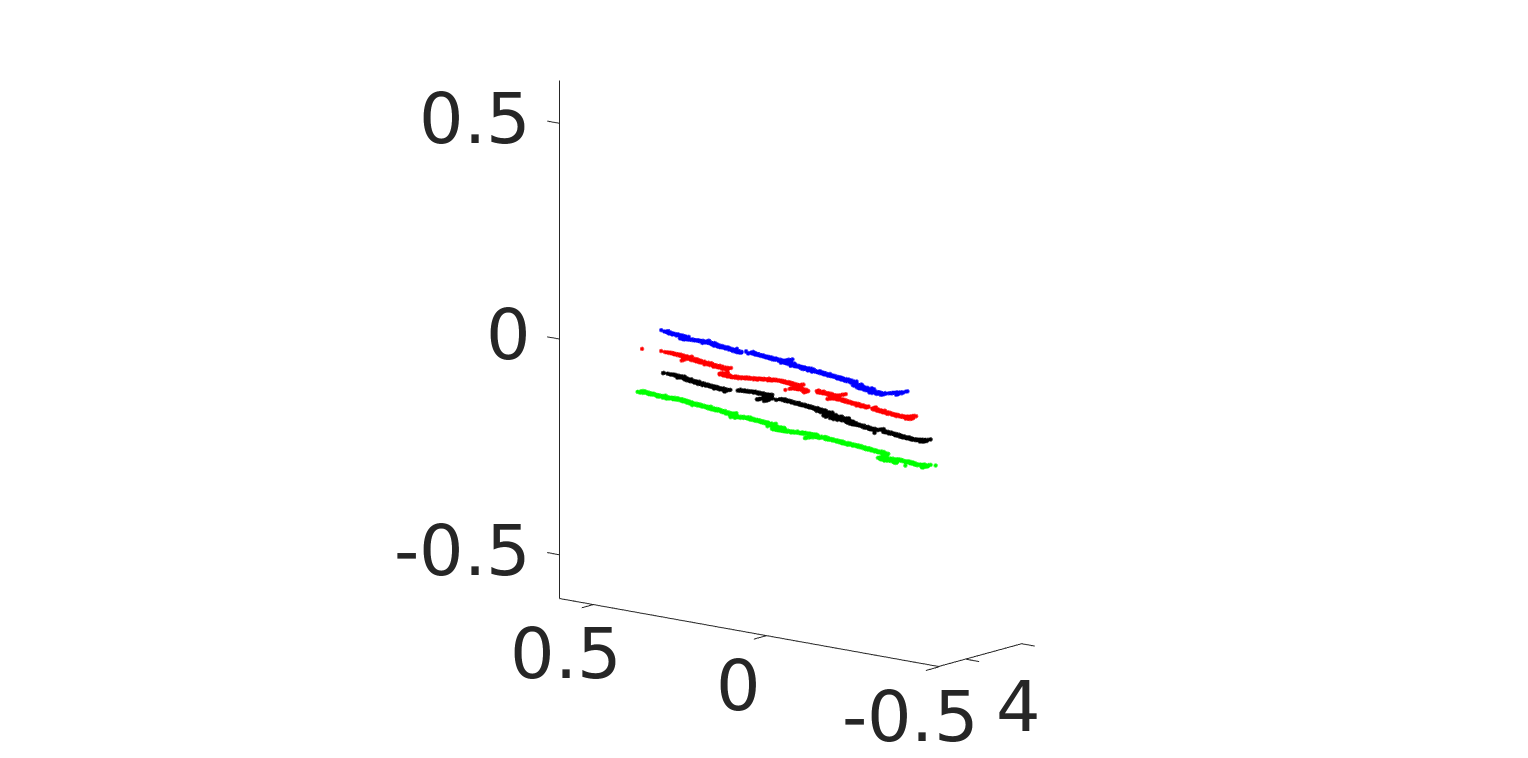}}\\
\hspace{8pt}%
\subfloat[]{%
    \label{fig:payload2D}%
\includegraphics[width=0.49\columnwidth, trim={1.5cm 0.5cm 3.5cm 2.1cm},clip]{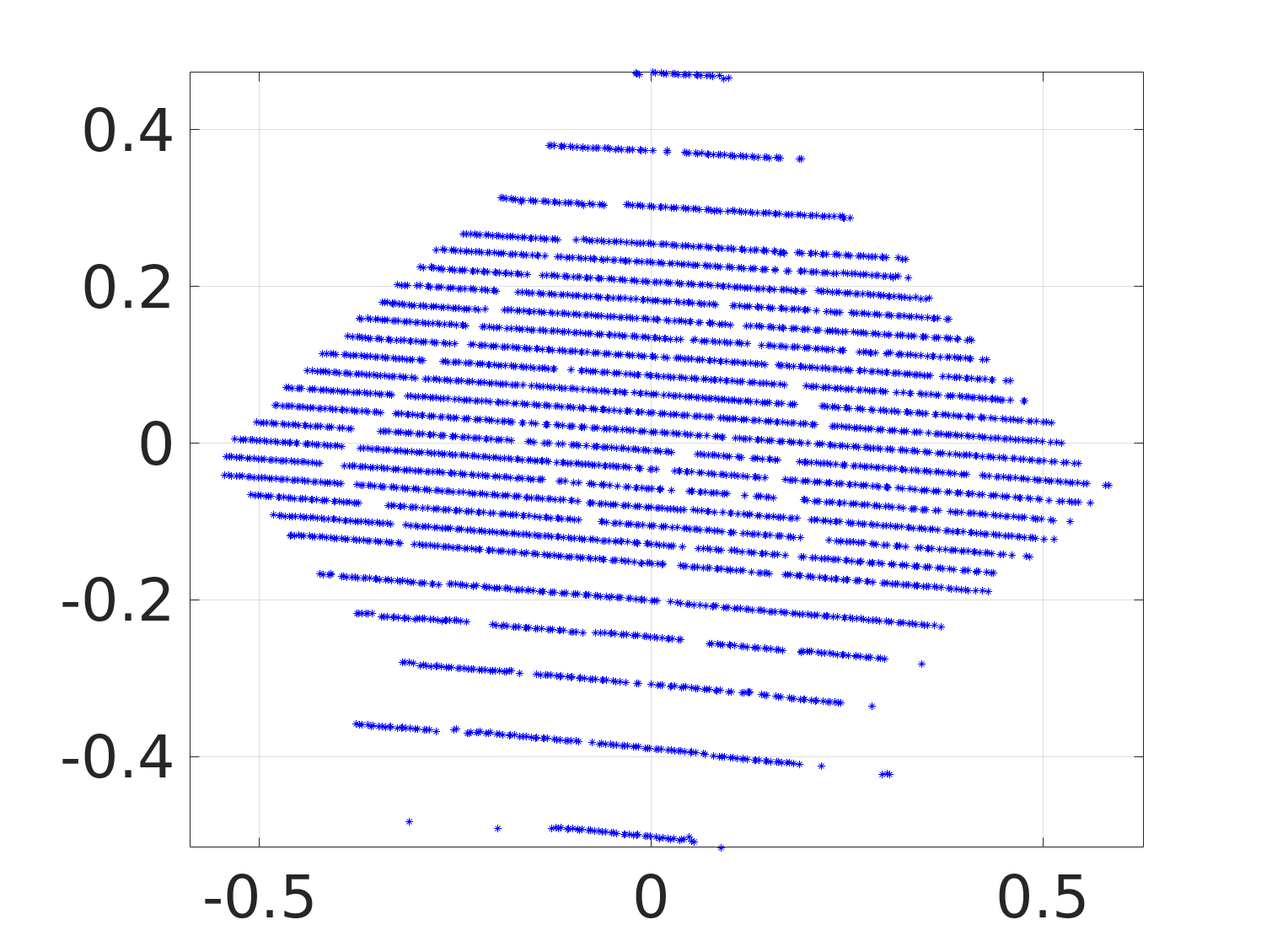}}%
\hspace{8pt}%
\subfloat[]{%
    \label{fig:payload_LN2D}%
\includegraphics[width=0.43\columnwidth, trim={.5cm 0 3cm 0},clip]{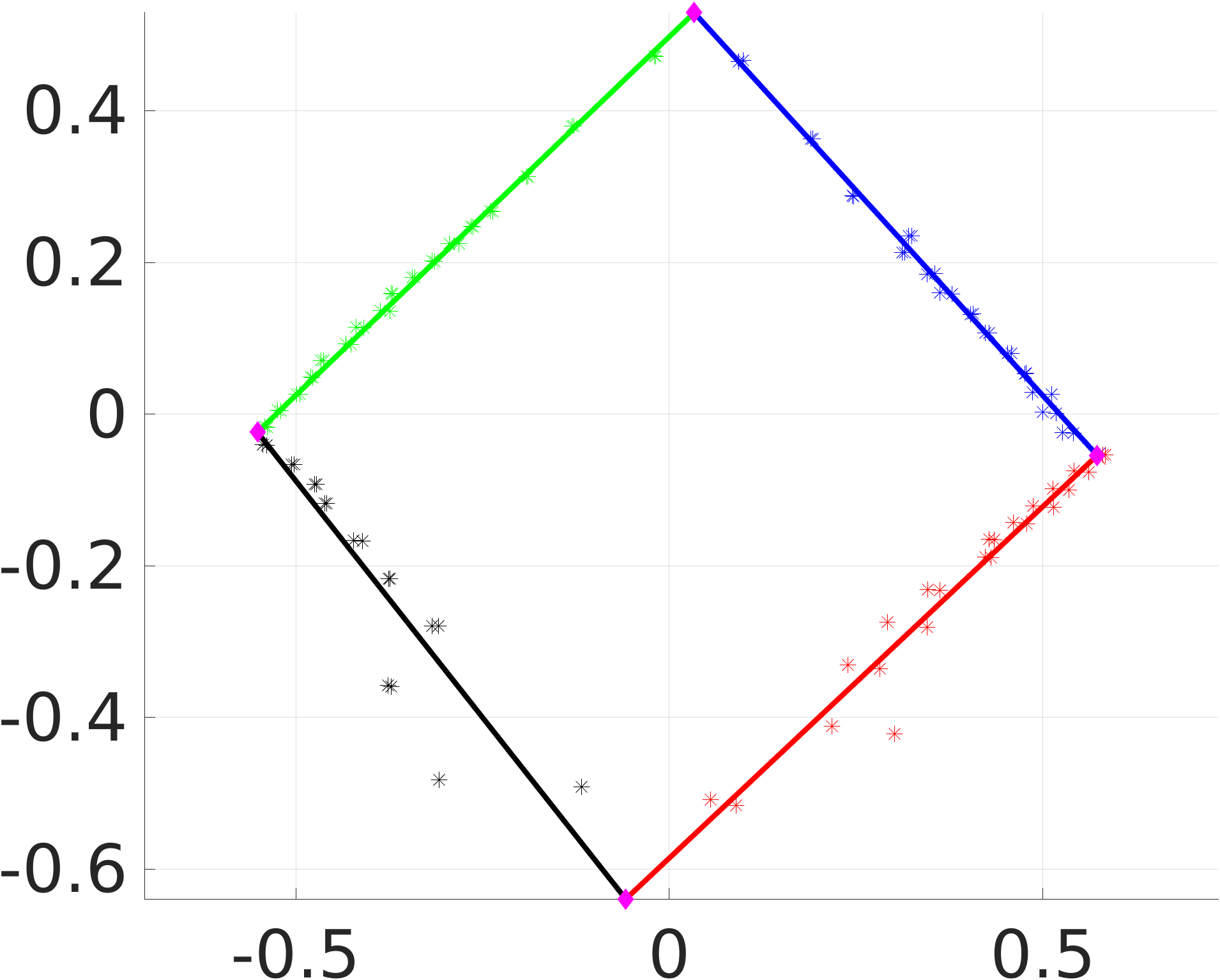}}\\
\hspace{8pt}%
\subfloat[]{%
\label{fig:reference_frame}%
\includegraphics[width=0.45\columnwidth, trim={0.05cm 0cm 0cm 1cm},clip]{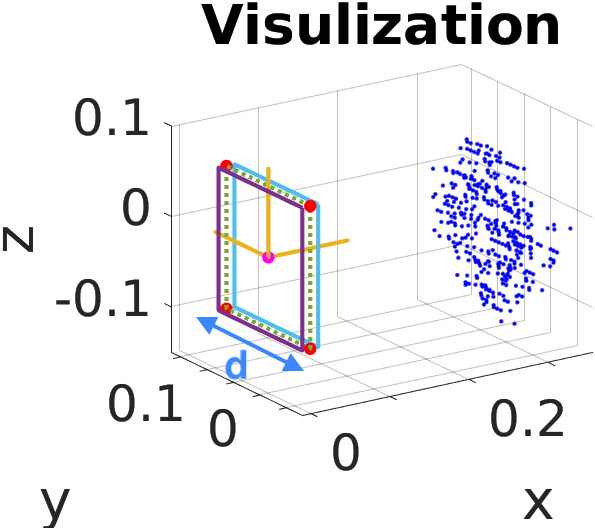}}%
\hspace{8pt}%
\subfloat[]{%
\label{fig:reference_frame_side}%
\includegraphics[width=0.45\columnwidth, trim={0.05cm 0cm 0cm 0.77cm},clip]{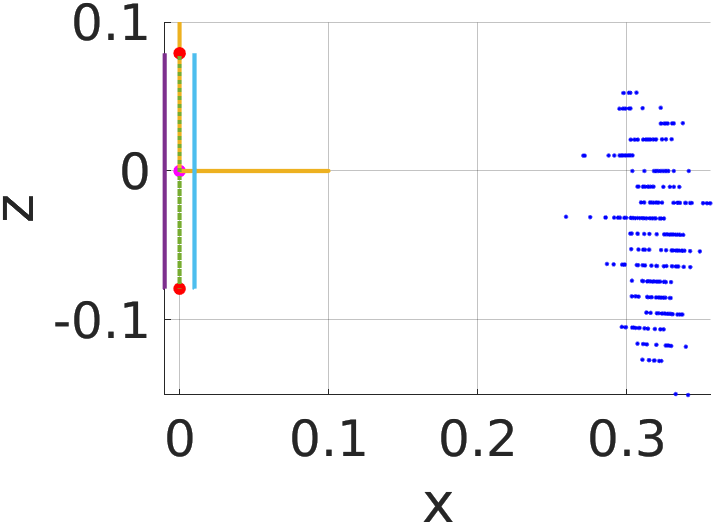}}%
\caption[] {Units are meters. \subref{fig:FigureLiDARScanExample} Twenty five LiDAR scans of a planar
    target. The point cloud is roughly 7~cm thick.
    \subref{fig:FigureLiDARScanExample3Rings} The ``noise'' in the point cloud is not
    random. A zoom for four of the rings (13, 15, 17, 19) is typical and shows
    systematic errors in distance. \subref{fig:payload2D} \lidar~points orthogonally
    projected to the plane defined by the normal. \subref{fig:payload_LN2D} Example
    edge points selected to regress a line via \textit{RANSAC}.
    \subref{fig:reference_frame} Target reference frame and real point cloud data.
    The dotted blue square is the reference target; its vertices are denoted
    $\{\bar{X}_i\}_{i=1}^4$. \subref{fig:reference_frame_side} A side-$(x-z)$-view
highlighting the non-zero thickness of a typical point cloud. These figures and all
others in the paper are of sufficient resolution that they can be blown up to see
detail.}%
\label{}%
\end{figure}

% \begin{figure}[b]
% 	\centering
% 	\includegraphics[width=\columnwidth]{graphics/FigureLiDARScanExample.png}
% 	\caption{Twenty five LiDAR scans of a planar target. The `noise' in the point cloud is not random.}
% 	\label{fig:FigureLiDARScanExample}
% \end{figure}
% The three times noise higher than the spec on the LiDAR points is per-ring based. 

Figure~\ref{fig:FigureLiDARScanExample} shows a 3D view of 25 scans from a
factory-calibrated \Velodyne on a diamond shaped planar
target. The zoom provided in Fig.~\ref{fig:FigureLiDARScanExample3Rings} shows that some of the rings are poorly calibrated, with a ``noise level'' that is three times higher than the manufacturer's specification. On average, the rings are close to the specification ($\pm$ five cm at less than 50 m) of the LiDAR, which means a maximum ten centimeters difference between points on the same ring. 

Systematic errors in the distance (or depth) measurement affect the estimation of the target's \textit{centroid}, which is
commonly used to determine the translation of the target with respect to the LiDAR, and the
target's \textit{normal vector}, which is used to define the plane of the target, as shown in
Fig.~\ref{fig:payload2D}. Subsequently, the point cloud is orthogonally projected to
this plane and the line boundaries of the target are found by performing
\textit{RANSAC} on the appropriate set of ring edges; see
Fig.~\ref{fig:payload_LN2D}. The lines along the target's boundaries then define its
vertices in the plane, which for later reference, we note are not constrained to be
compatible with the target's geometry. 

Once the vertices in the plane of the target have been determined, then knowledge of
the target's normal vector allows the vertices to be lifted back to the coordinates of the
point cloud. This process may be repeated for multiple scans of a target, aggregates
of multiple scans of a target, or several targets, leading to a list of target
vertices $\{ X_i \}_{i=1}^{4n} $, where $n$ is the number of target poses.

The target is typically designed so that the vertices are easily distinguishable in
the camera image. Denoting their corresponding coordinates in the image plane by $\{
Y_i \}_{i=1}^{4n}$ completes the data required for the conceptual fitting problem in
\eqref{eq:ConceptualProblem}. While the cost to be minimized is nonlinear and non-convex, this is
typically not a problem because CAD data can provide an adequate initial guess for
local solvers, such as Levenberg-Marquardt; see \cite{zhou2018automatic} for example.

% \begin{figure}[b]
% \vspace{-2mm}
% 	\centering
% 	\includegraphics[width=\columnwidth]{graphics/FigureLiDARScanExample3Rings.png}
% 	\caption{[Should go side by side with previous figure] Zoom in on three of the rings (7, 12, and 17) of Fig.~\fig{fig:FigureLiDARScanExample} }
% 	\label{fig:FigureLiDARScanExample3Rings}
% \squeezeup \squeezeup
% \end{figure}

\subsection{Our Contributions}
\label{sec:Contributions}
Our contributions can be summarized as follows. 
\begin{enumerate}
\setlength{\itemsep}{.15in}
\renewcommand{\labelenumi}{(\alph{enumi})}
% 	\item We make use of the target's geometry when estimating its vertices. In
% particular, if the target is a diamond, vertices form right angles and the
% lengths of the sides are equal. We show that this simple change improves the
% accuracy of estimating the vertices. Essentially, the extra constraints allow
% all of the target's boundary points to be pulled into the regression problem at
% once, instead of parsing them into individual edges of the target, where small
% numbers of points on some of the edges will accentuate the quantization effects due to sparsity
% in a LiDAR point cloud. \jkh{remove edge parsing and line-ransac}

% \item We make use of the target's geometry when estimating its vertices. In
% particular, if the target is a diamond, vertices form right angles and the
% lengths of the sides are equal. 

\item We introduce a novel method for estimating the rigid-body transform from target
    to LiDAR, $H_{T}^{L}$. For the point cloud pulled back to the origin of the LiDAR
    frame via the current estimate of the transformation, the cost is defined as the
    sum of the distance of a point to a 3D-reference target of known
    geometry\footnote{It has been given non-zero volume.} in the LiDAR frame, for
    those points landing outside of the reference target, and zero otherwise. We use
    an $L_1$-inspired norm in this work to define distance. Consistent with
    \cite{liao2018extrinsic}, we find that this is less sensitive to outliers than
    $L_2$-line fitting to edge points\footnote{Defined as the left-right end points
    of each LiDAR ring landing on the target.} using \textit{RANSAC}
    \cite{zhou2018automatic}.

\item In the above, by using the entire target point cloud when estimating the
    vertices, we avoid all together the delicate task of identifying the edge points
    and parsing them into individual edges of the target, where small numbers of
    points on some of the edges accentuate the quantization effects due to sparsity
    in a LiDAR point cloud.

% 	\item We exploit the fact that the variance of the camera data is
% significantly lower than that of the LiDAR data. Hence, after the LiDAR to
% camera transformation has been estimated by solving the PnP problem, we
% introduce a refinement step to the estimated pose of the LiDAR target given the
% current estimate of the LiDAR to camera transformation. In this process, the
% modified vertices remain consistent with the target geometry and the target's
% point cloud. \jwg{Called Ridiculous in the reviews}

\item We provide a round-robin validation study to compare our approach to a
    state-of-the-art method, namely \cite{zhou2018automatic}. A novel
    feature of our validation study is the use of the camera image as a proxy for ground truth; in the
    context of 3D semantic mapping, this makes sense.  In addition to a standard PnP method~\cite{lepetit2009epnp} for
    estimating the rigid-body transformation
    from LiDAR to camera, we also investigate maximizing intersection over union (IoU) to estimate the rigid-body transformation. Our algorithms are validated on various numbers of targets.

\item Code for our method, our implementation of the baseline, and all the
    data sets used in this paper are released as
    \href{https://github.com/UMich-BipedLab/extrinsic_lidar_camera_calibration}{open
    source}; see \cite{githubFile}.
\end{enumerate}

\begin{comment}

\subsection{More on Related Work}
\label{sec:RelatedWork}

Papers to talk about

\begin{itemize}
    \item Simultaneous intrinsic and extrinsic
    \item When there is a large separation between lidar and camera
    \item SVD prevleance
    \item Ransac
    \item \textbf{Most closely related to our work:} Extrinsic calibration of lidar and camera with polygon
\end{itemize}
\end{comment}

% \subsection{Paper Outline}
% The remainder of the paper is structured....

%\input{Related}
\section{Finding the LiDAR Target Vertices}
\label{sec:LiDARTargetVertices}

We assume that each target is planar,
square, and rotated in the frame of the LiDAR by roughly 45$^\circ$ to form a diamond
as in Fig.~\ref{fig:FigureLiDARScanExample}. As indicated in
\cite{liao2018extrinsic,zhou2018automatic}, placing the targets so that the rings of
the LiDAR run parallel to its edges leads to ambiguity in the vertical position due
to the spacing of the rings.  We assume that standard methods have been used to
automatically isolate the target's point cloud \cite{huang2019lidartag} and speak no
further about it. 

We take as a target's features its four vertices, with their coordinates in the LiDAR
frame denoted $\{ X_i \}_{i=1}^{4};$ when useful, we abuse notation and pass from
ordinary coordinates to homogeneous coordinates without noting the distinction. The
vertices  $\{ X_i \}_{i=1}^{4} $ are of course not directly observable in the point
cloud. This section will provide a new method for estimating their coordinates in the
frame of the LiDAR using an $L_1$-like norm. 

%It will also
%provide a simple improvement to existing methods that use least squares fitting.

% \begin{figure}[t]%
% \centering
% \subfloat[]{%
% \label{fig:reference_frame}%
% \includegraphics[width=0.38\columnwidth, trim={0.05cm 0cm 0cm 1cm},clip]{reference_frame.png}}%
% \hspace{1pt}%
% \subfloat[]{%
% \label{fig:reference_frame_side}%
% \includegraphics[width=0.38\columnwidth, trim={0.05cm 0cm 0cm 0.77cm},clip]{reference_frame_side.png}}%
% \caption[]{Target reference frame and real point cloud
% data. \subref{fig:reference_frame}  The dotted blue square is the reference target; its vertices are denoted $\{\bar{X}_i\}_{i=1}^4$. \subref{fig:reference_frame_side} A
% side-$(x-y)$-view highlighting the non-zero thickness of a typical point cloud.}%
% \label{fig:}%
% \squeezeup
% \end{figure}

 %trim={<left> <lower> <right> <upper>}
\begin{figure}[t]%
\centering
\subfloat[]{%
    \label{fig:disturbance}%
    \centering
\includegraphics[width=0.48\columnwidth, trim={3cm 0.63cm 4.5cm 3cm},clip]{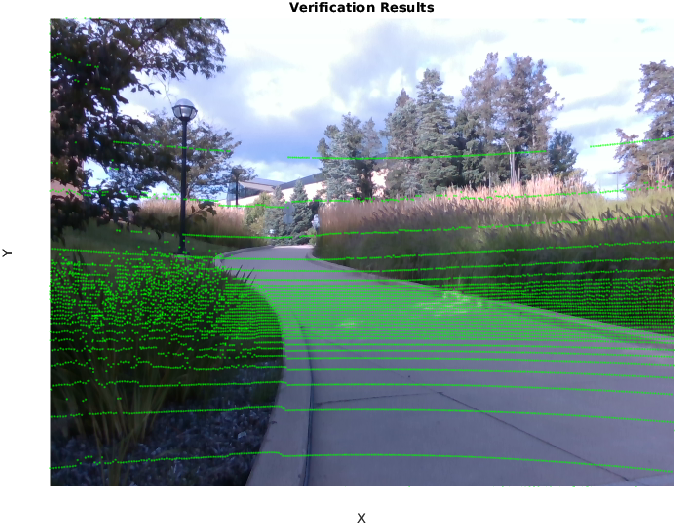}}%
\hspace{10pt}%
\subfloat[]{%
\label{fig:undisturbance}%
\centering
\includegraphics[width=0.46\columnwidth, trim={3cm 0.6cm 4.5cm 3cm},clip]{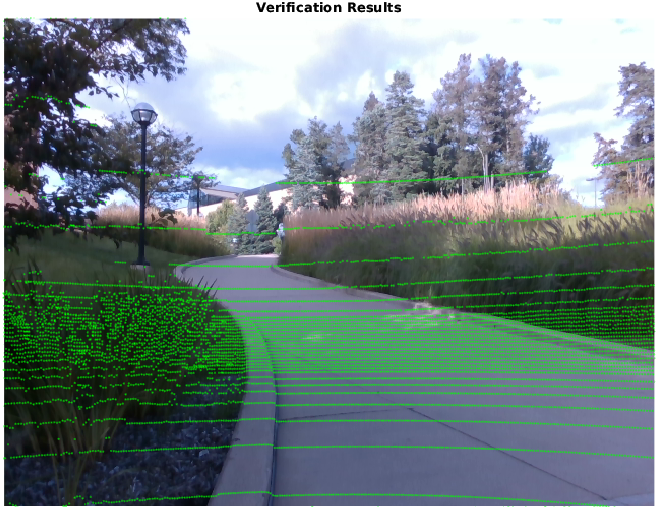}}%
\caption[]{\subref{fig:disturbance} shows that a calibration result is not usable if
it has a few degrees of rotation error and a few percent of translation error.
\subref{fig:undisturbance} shows good alignment of a \lidar~point cloud projected
onto a camera image.}% \squeezeup
\end{figure}

\subsection{Remarks on LiDAR Point Clouds}
\label{sec:lidar_analysis} 
\lidars have dense regions and sparse regions along the z-axis, as shown in
Fig.~\ref{fig:payload2D}. For a 32-beam Velodyne Ultra Puck, we estimate the
resolution along the z-axis is $0.33^\circ$ and $1.36^\circ$ in the dense and sparse
regions, respectively. A point's distance resolution along the y-axis is roughly
$0.1^\circ$. 
The quantization error could be roughly computed from: $d\sin{\theta}$
when a point is at $d$ meter away. As a result, the quantization error could get
quite large if one place the tag at a far distance. Figure~\ref{fig:disturbance}
shows that a $1^\circ$ of rotation error on each axis and 5\% error in translation
can significantly degrade a calibration result. As noted in
\cite{liao2018extrinsic,zhou2018automatic}, it is essential to place a target at an
appropriate angle so that known geometry can mitigate quantization error in the $y$-
and $z$-axes. 

%% related to refinement
% \jkh{please check}
% To overcome quantization error in the $y$-axis and $z$-axes, we proposed a refinement step in
% \cite{huang2019improvements} to modify the estimated vertices. The refinement step conditions
% on the current estimate of the projection map, image corners as well as a weighted
% loss on \eqref{eq:JKHcost}. However, due to the limited space, we do not apply this step.
% To overcome quantization error in the $x$-axis, we accumulate a few
% scans to estimate a target's pose and ``condition'' on the current estimate of LiDAR
% to image plane projection matrix to refine the target's pose. 

\subsection{New Method for Determining Target Vertices}
\label{sec:NewMethod}

\begin{figure}[t]
\centering
\includegraphics[width=1\columnwidth, bb=0 0 1892 1458]{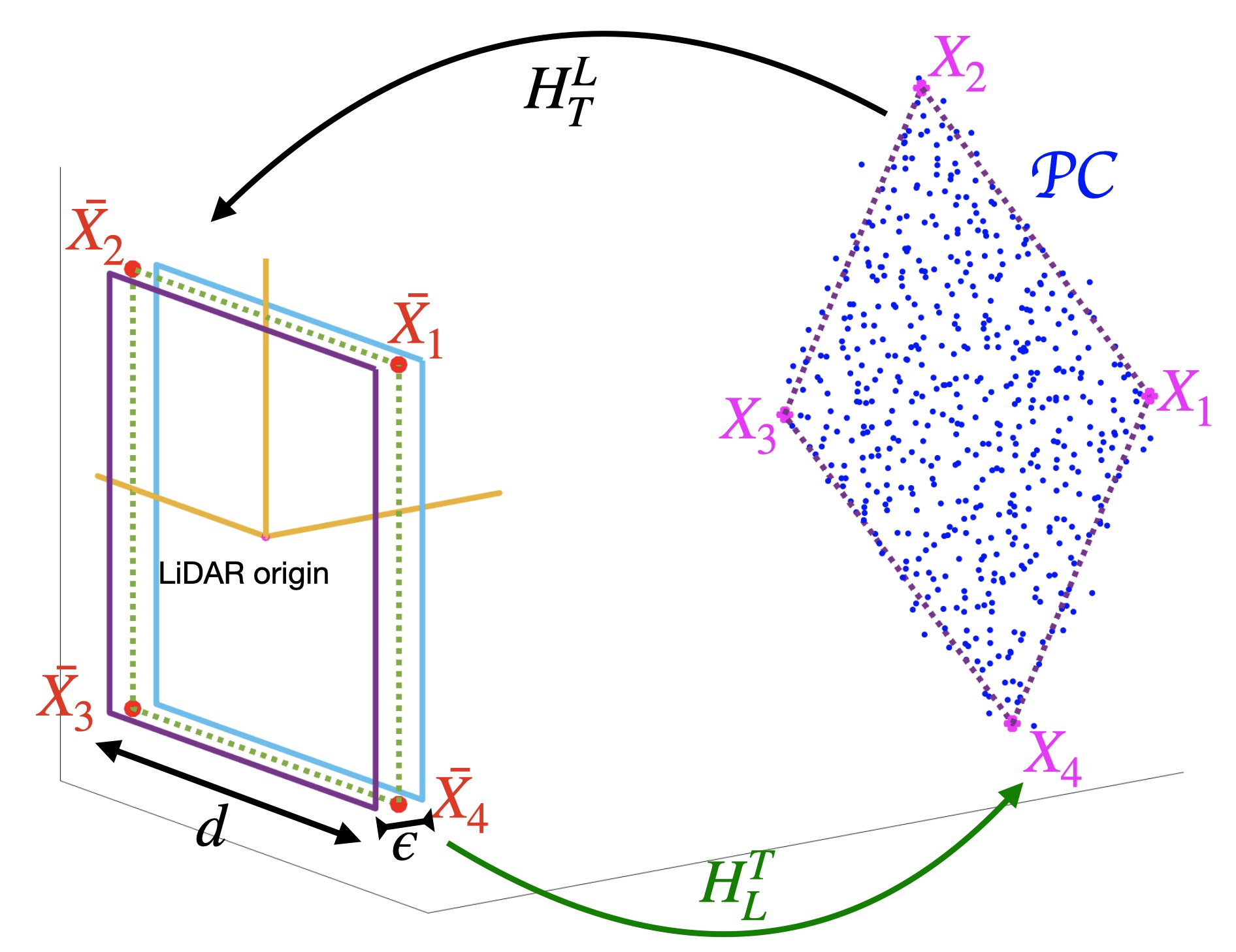}
\caption{This conceptual figure illustrates the proposed method to estimate \lidar vertices. The target's reference in the \lidar frame is defined by ($\bar{X_1}, \cdots, \bar{X_4}$) with depth $\epsilon$.
The rigid-body transformation $H_T^{L}$ (black arrow) pulls back the target's $\pc$ to a target reference about the \lidar origin. The actual vertices ($X_1, \cdots, X_4$) of the target are estimated by \eqref{eq:vertices}, using the inverse transformation $H_L^{T}$ (green arrow).}
\label{fig:FormulaIllustraion}
\end{figure}

Let ${\cal PC}$ denote the LiDAR point cloud of the target and let the collection of 3D points be
${\cal X}_i$ so that ${\cal PC} = \{ {\cal X}_i\}_{i=1}^N$, where $N$ is the number of points on the target. For the extrinsic calibration problem, we need to estimate the target vertices in the LiDAR
frame. As in Fig.~\ref{fig:FormulaIllustraion}, let $H_L^T$
be the rigid-body transformation from a reference target in the \lidar frame with vertices $\{ \bar{X}_i \}_{i=1}^4$, onto the point cloud. We use the inverse transform $H_T^L:=(H_L^T)^{-1}$  to pull back the target's point cloud to the origin of the \lidar and measure the error there.

% \begin{remark}
%     \label{remark:Xis}
%     To be clear, what we seek is a ``best estimate'' of the target vertices in the LiDAR
%     frame and not the transformation itself. Our method is indirect because from the
%     point cloud, we estimate a ``best'' LiDAR to target transformation, call it
%     $H_L^{T^*}$, and use it to define the vertices 

%     \begin{equation}
%     \label{eq:vertices}
%     X_i^*:=H_L^{T^*}(\bar{X}_i), ~~1 \le i \le 4.
%     \end{equation}

%     The correspondences of the estimated vertices with the physical top, bottom, and left
%     or right sides of the target are not needed at this point.  $\blacksquare$
% \end{remark}

For $a \ge 0$ and $\lambda \in \real$, define 
% \jwg{Please check again $\chi$ vs ${\cal X}$}
\begin{equation}
    \label{eq:L1cost}
    c(\lambda,a):=\begin{cases} 
    \min\{ |\lambda-a|, |\lambda + a| \} & \text{if}~|\lambda| >a \\
    0 & \text{otherwise}
    \end{cases};
\end{equation}
see also the ``soft $L_1$ norm'' in \cite{liao2018extrinsic}. Let
$\{\tilde{{\cal X}}_i\}_{i=1}^N:=H_T^L(\mathcal{PC}):=\{ H_T^L({\cal X}_i) \}_{i=1}^N$  denote
the pullback of the point cloud by $H_T^L$, and denote a point's Cartesian coordinates by
$(\tilde{x}_i,\tilde{y}_i,\tilde{z}_i)$. The cost is defined as 
\begin{equation}
\label{eq:JKHcost}
    C(H_T^L(\mathcal{PC})):=\sum_{i=1}^{N} c(\tilde{x}_i,\epsilon) +
    c(\tilde{y}_i,d/2) +  c(\tilde{z}_i,d/2),
\end{equation}
where $d$ is determined by the size of the (square) target, and the only tuning parameter is $\epsilon > 0$, the thickness of the ideal target. The value of $\epsilon$ is based on the standard deviation of a target's return points to account for the noise level of the depth measurement; see Fig.~\ref{fig:reference_frame_side}.

We propose to determine the optimal rigid-body transformation, with rotation matrix $R_T^L$ and translation vector $T_T^L$,  by
\begin{equation}
\label{eq:OurNewCost}
    H_T^{L^*} := \argmin_{R_T^L, T_T^L}
   C(H_T^L(\mathcal{PC})),
\end{equation}
and to define the estimated target vertices by
    \begin{equation}
    \label{eq:vertices}
    X_i^*:=H_L^{T^*}(\bar{X}_i), ~~1 \le i \le 4.
    \end{equation}

\begin{remark} It is emphasized that the target vertices are being determined without
extraction of edge points and their assignment to a side of the target.  The correspondences of the estimated vertices with the physical top, bottom, and left or right sides of the target are not needed at this point.  
\end{remark}

\begin{remark}
    \label{remark:Xis}
    The cost in
\eqref{eq:JKHcost} treats the target as a rectangular volume. 
    To be clear, what we seek is a ``best estimate'' of the target vertices in the LiDAR
    frame and not the transformation itself. Our method is indirect because from the
    point cloud, we estimate a ``best'' LiDAR to target transformation, $H_L^{T^*}$, and use it to define the vertices by \eqref{eq:vertices}.
\end{remark}

\section{Image Plane Corners and Correspondences with the LiDAR Vertices}
\label{sec:CameraStuff}

For a given camera image of a LiDAR target, let $\{ \pre[C]Y_i\}_{i=1}^4$ denote the corners of the camera image. We assume that these have been obtained through the
user's preferred method, such as corner detection \cite{harris1988combined,
rosten2006machine, rosten2008faster}, edge detection \cite{canny1987computational,
lim1990two, vincent2009descriptive}, or even manual selection. The resulting camera corners are used in both the proposed method and the baseline. This is not the hard
part of the calibration problem. To achieve simple correspondences $X_i
\leftrightarrow \pre[C]Y_i$, the order of the indices of $\{ X_i\}_{i=1}^4$ may need
to be permuted; we use the vertical and horizontal positions to sort them; see \cite{githubFile}.

Once we have the correspondences, the next step is to project the vertices of the
\lidar~target, $\begin{bmatrix} x_i & y_i & z_i & 1 \end{bmatrix}^\transpose = X_i$,
into the image coordinates. The standard relations are \cite{hartley2000multiple,
forsyth2002computer}

\begin{align}
% \begin{equation}
    \label{eq:projection_linear}
    % \left(R_L^C^*, T_L^C^*\right) &=  \argmin_{R_L^C,
    % T_L^C}\sum_{i=1}^{N}\norm{\Pi_K\left(\pre[L]Y_i; R_L^C, T_L^C\right)-\pre[C]Y_i}_2^2\\
    %         &= \argmin_{R_L^C,
    %         T_L^C}\norm{\Pi_K\left(\pre[\textbf{L}]\textbf{Y}; R_L^C, T_L^C\right) - 
    %     \pre[\textbf{C}]\textbf{Y}}_F^2,
    \begin{bmatrix}
        u^\prime\\
        v^\prime\\
        w^\prime
    \end{bmatrix} &=
    \begin{bmatrix}
        f_x & s & c_x \\
        0 & f_y & c_y \\
        0 & 0 & 1
    \end{bmatrix}
    \begin{bmatrix}
    \mathbb{1}_{3\times3} \\ \zeros[1\times3]
    \end{bmatrix}^\transpose
    \begin{bmatrix}
        R_L^C & T_L^C \\
        \zeros[1\times 3] & 1 \\
    \end{bmatrix}
    \begin{bmatrix}
        x_i \\
        y_i \\
        z_i \\
        1
    \end{bmatrix}\\
% \end{equation}
% \begin{align}
% \begin{equation}
    \label{eq:projection_nonlinear}
    \pre[L]Y_i &= \begin{bmatrix}
        u & v & 1
    \end{bmatrix}^{\transpose} = 
    \begin{bmatrix}
        \frac{u^\prime}{w^\prime} & \frac{v^\prime}{w^\prime} & 1
    \end{bmatrix}^{\transpose},
% \end{equation}
\end{align}
where \eqref{eq:projection_linear} includes the camera's intrinsic parameters and the
extrinsic parameters $(R_L^C, T_L^C)$ that we seek. 

For later use, we combine \eqref{eq:projection_linear} and \eqref{eq:projection_nonlinear} to define 
\begin{equation}
\label{eq:Projection}
\Pi\left(X_i; R_L^C, T_L^C \right):=\pre[L]Y_i, 
\end{equation}
the projection map from LiDAR coordinates to image coordinates. Note that it is
a function of both the extrinsic variables and the LiDAR vertices.

%\input{ProblemStatementAndExtrinsicCalibration}
%\input{ProblemStatementANDLiDARTagPose}
% \section{Extrinsic Optimization and Pose Refinement}
\section{Extrinsic Transformation Optimization}
\label{sec:extrinsic_opt}
This section assumes the vertices of the target in the LiDAR frame and in the
camera's image plane have been determined, along with their correspondences. The
optimization for the extrinsic transformation can be formulated in a standard PnP
problem: minimize Euclidean distance of the corresponding corners. We also propose 
maximizing the intersection over union (IoU) of the corresponding projected polygons.

% This section assumes the vertices of the target in the LiDAR frame and in the
% camera's image plane have been determined, along with their correspondences. The
% optimization for the extrinsic transformation can be formulated in two ways: minimize
% Euclidean distance of the corresponding corners or maximize the
% intersection over union (IoU) of the corresponding polygons. 

\subsection{Euclidean distance}
The standard PnP formulation is

\begin{align}
\label{eq:PnP}
    \left({R_L^C}^*, {T_L^C}^*\right) &:=  \argmin_{R,
    T}\sum_{i=1}^{4n}\norm{\Pi\left(X_i; R, T\right)-\pre[C]Y_i}_2^2  \nonumber \\
                                  &=\argmin_{R,
                                  T}\sum_{i=1}^{4n}\norm{\pre[L]Y_i-\pre[C]Y_i}_2^2,
\end{align}
where $\pre[C]Y_i\in \realnumbers^2$ are the camera corners, $\pre[L]Y_i\in
\realnumbers^2$ are defined in \eqref{eq:Projection}, and $n$ is the number of target
poses.

% The second method requires some geometry background. Assume we have a set of N
% vertices of a polygon $(x_i, y_i)$ in Cartesian coordinates. To sort the vertices in
% counterclockwise (CCW) direction, we apply Graham scan, a method to find the
% convex hull of a finite set of points on a plane \red{cite}. Given a set of sorted
% CCW points of vertices of a polygon $(x_i, y_i)$, the area of the polygon $(\A)$ can be calculated via
% Shoelace algorithm \red{cite}:
% \begin{equation}
%     \label{eq:Shoelace}
%     \A = \frac{1}{2}\left|\sum_{i=1}^{N} \det\begin{bmatrix} x_i & x_{i+1} \\ y_i &
%     y_{i+1}\end{bmatrix}\right|,
% \end{equation}
% where $x_{n+1} = x_1, x_0 = x_n$ as well as $y_{n+1} = y_1, y_0 = y_n$.

\subsection{IoU optimization}
\label{sec:IoU}
% To compute the IoU of two projected polygons of the targets in an image plane, we
% define the intersection as a polygon with known coordinates\footnote{The vertices of
% the polygon consist of the 2D corners and the 2D line intersections and thus can be
% computed efficiently.}. We sort the vertices 
% ($\pre[L]Y_i, \text{ and }\pre[C]Y_i$)
% of the polygon counterclockwise using
% Graham scan, a method to find the convex hull of a finite set of points on a plane
% \cite{graham1972efficient, andrew1979another}. The area of each polygon is easily computed. The area of their union is calculated via the Shoelace algorithm \bh{\cite[eq. (??)]{braden1986surveyor}.}
\begin{figure}[t]%
\centering
\subfloat[]{%
    \label{fig:TwoPolygons}%
    \centering
\includegraphics[width=0.46\columnwidth]{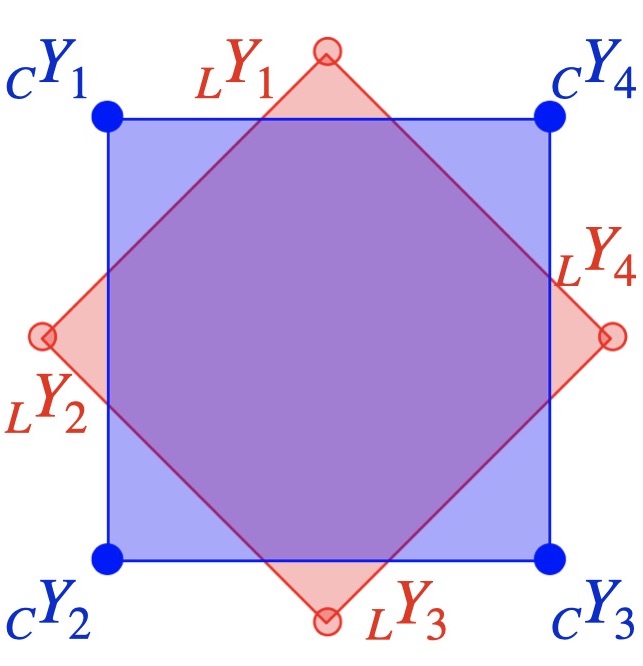}}%
\hspace{10pt}%
\subfloat[]{%
\label{fig:IntersectionPolygon}%
\centering
\includegraphics[width=0.46\columnwidth]{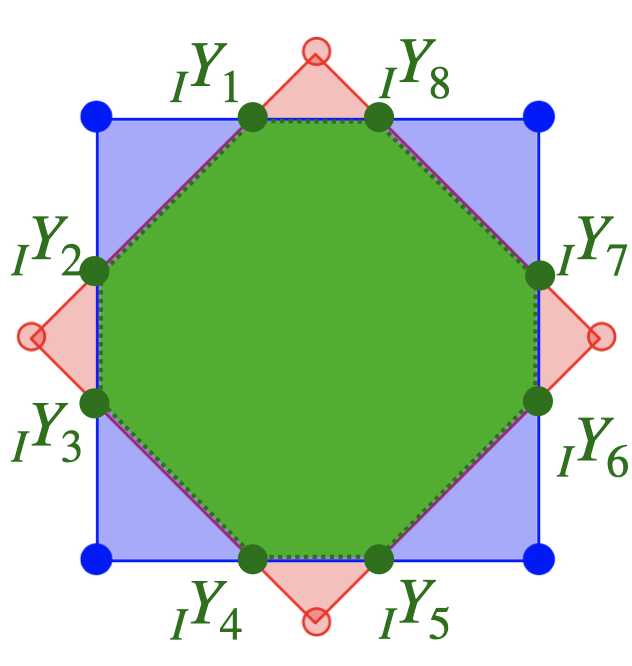}}%
\caption[]{Two polygons in an image plane. \subref{fig:TwoPolygons} the red polygon  $\pre[L]\vx =\{\pre[L]Y_i\}_{i=1}^{4}$ represents the estimated vertices projected from the \lidar frame to the image plane, while the blue polygon $\pre[C]\vx =\{\pre[C]Y_i\}_{i=1}^{4}$ represents the vertices of the actual camera image of the target.
\subref{fig:IntersectionPolygon} shows the intersection (marked in green) of the two polygons, with the vertices labeled as $\pre[I]\vx =\{\pre[I]Y_i\}_{i=1}^{8}$.}%
\end{figure}
For a given polygon, let $\vx := \{Y_i | Y_i=: (x_i, ~y_i)\}_{i=1}^N$ be the coordinates of the $N$ vertices, ordered counterclockwise. The area of the polygon is computed via the Shoelace algorithm \cite[eq. (3.1)]{ochilbek2018new} \cite{braden1986surveyor},
\begin{equation}
    \label{eq:Shoelace}
    \A(\vx) = \A(Y_1, \cdots, Y_N) = \frac{1}{2}\left|\sum_{i=1}^{N} \det\begin{bmatrix} x_i & x_{i+1} \\ y_i &
    y_{i+1}\end{bmatrix}\right|,
\end{equation}
where $x_{N+1} := x_1$ and $y_{N+1} := y_1$. If $\vx$ is empty, the area is taken as zero. We now apply this basic area formula to propose an IoU-based cost function for extrinsic calibration.

Let $\pre[L]\vx:=\{\pre[L]Y_i\}_{i=1}^{4n}$  be the vertices of the estimated target polygons projected from the \lidar frame to the camera frame as in \eqref{eq:Projection}, and let $\pre[C]\vx:=\{\pre[C]Y_i\}_{i=1}^{4n}$ be the vertices of the corresponding camera images of the targets, as in Fig.~\ref{fig:TwoPolygons}. If their intersection is nonempty, we define it as a polygon with known coordinates\footnote{The vertices of
the polygon consist of the 2D corners and the 2D line intersections and thus can be computed efficiently; see Fig.~\ref{fig:IntersectionPolygon}}, denoted as $\pre[I]\vx := \{\pre[I]Y_i\}_{i=1}^M$, where $M\ge 0$ is the number of intersection points; see Fig.~\ref{fig:IntersectionPolygon}. We sort the three sets of vertices of the polygons counterclockwise using Graham's scan algorithm, a method to find the convex hull of a finite set of points in a plane \cite{graham1972efficient, andrew1979another}. The IoU of the two polygons is then
% \begin{equation}
%     \label{eq:IoUCost}
%     \resizebox{0.95\hsize}{!}{%
%     $IoU(\pre[L]\vx,\pre[C]\vx) = \frac{\A(\{\pre[I]Y_i\}_{i=1}^M)}{\A(\{\pre[L]Y_i\}_{i=1}^{4n}) + \A(\{\pre[C]Y_i\}_{i=1}^{4n}) - \A(\{\pre[I]Y_i\}_{i=1}^M)}$}.
% \end{equation}
\begin{equation}
    \label{eq:IoUCost}
    IoU(\pre[L]\vx,\pre[C]\vx,\pre[I]\vx) = \frac{\A(\pre[I]\vx)}{\A(\pre[L]\vx) + \A(\pre[C]\vx) - \A(\pre[I]\vx)}.
\end{equation}
The resulting optimization problem is
\begin{align}
\label{eq:IoUOptimization}
    \left({R_L^C}^*, {T_L^C}^*\right) &:= \argmax_{R,T}IoU(\pre[L]\vx,\pre[C]\vx)\\
                                       &= \argmax_{R,T}IoU(\Pi\left(\{X_i\}_{i=1}^{4n}; R, T\right),\pre[C]\vx), \nonumber
\end{align}
where \eqref{eq:Projection} has been used to make the dependence on the rigid-body transformation explicit.

\section{Experimental Results}
\label{sec:ExperimentsAndResults}
In this section, we extensively evaluate our proposed method on seven different
scenes through a form of ``cross-validation'': in a round-robin fashion, we estimate an extrinsic transformation using data from
one or more scenes and then evaluate it on the remaining scenes. The quantitative
evaluation consists of computing pixel error per corner, where we take the image
corners as ground truth. We also show qualitative validation results by projecting
the LiDAR scans onto camera images; we include here as many as space allows, with
more scenes and larger images available at \cite{githubFile}. 

\subsection{Data Collection}
\label{sec:DataGathering}
The scenes include both outdoor and indoor settings. Each scene includes two targets,
one approximately 80.5~cm square and the other approximately 15.8~cm square, with the
smaller target placed closer to the camera-LiDAR pair. We use an \textit{Intel
RealSense Depth Camera D435} and a \textit{32-Beam Velodyne ULTRA Puck LiDAR},
mounted on an in-house designed torso for a Cassie-series bipedal robot
\cite{CassieAutonomy2019ExtendedEdition}. From the CAD file, the camera is roughly
20~cm below the LiDAR and 10~cm in front of it. The angle of the camera is
adjustable. Here, its ``pitch'', in the LiDAR frame, is approximately zero. 

A scan consists of the points collected in one revolution of the LiDAR's 32 beams.
The data corresponding to a single beam is also called a ring. For each scene, we
collect approximately 10~s of synchronized data, resulting in approximately 100 pairs
of scans and images. 

For each target, five consecutive pairs of LiDAR scans and camera images are selected
as a data set. For each data set, we apply two methods to estimate the vertices of
the targets, a baseline and the method in Sec.~\ref{sec:NewMethod}. We then apply both PnP and IoU optimizations to find the rigid-body transformation, see Sec.~\ref{sec:extrinsic_opt}.
% For each
%  \jwg{should have been commented out: target, 25 consecutive pairs of LiDAR scans and camera images are selected and
% divided into 5 data sets.} 
% For each data set, we apply two methods to estimate the
% vertices of the targets, a baseline and the method in Sect.~\ref{sec:NewMethod}.

\subsection{Baseline Implementation for LiDAR Vertices}
\label{sec:Baseline}
As a baseline, we use the method in \cite{zhou2018automatic}. Because an open-source implementation was not released, we built our own, attempting to be as faithful as possible to the described method. 

For each scan, the large and small targets are individually extracted from background
\cite{huang2019lidartag}. For each target and group of five scans, we compute the
extracted point cloud's centroid and center the point cloud about the origin. SVD is
then used to find the target's normal and to orthogonally project the point cloud
onto the plane defined by the normal and the centroid. For each scan, and for each
ring hitting the target, the left and right end points of the ring are selected and
then associated with one of the four edges of the target. Lines are fitted to each
collection of edge points using least-squares and \textit{RANSAC}. The vertices are
obtained as the intersections of the lines in the plane defined by the target's
normal and centroid. The four vertices are then re-projected to 3D space, as in
Fig.~\ref{fig:payload_LN2D}.

\subsection{Camera Corners and Associations}
\label{sec:CameraCorners}
\begin{figure}[t]%
\centering
\subfloat[]{%
\label{fig:Cannyedges}%
\includegraphics[width=0.48\columnwidth, trim={3.5cm 12.6cm 3cm 10cm},clip]{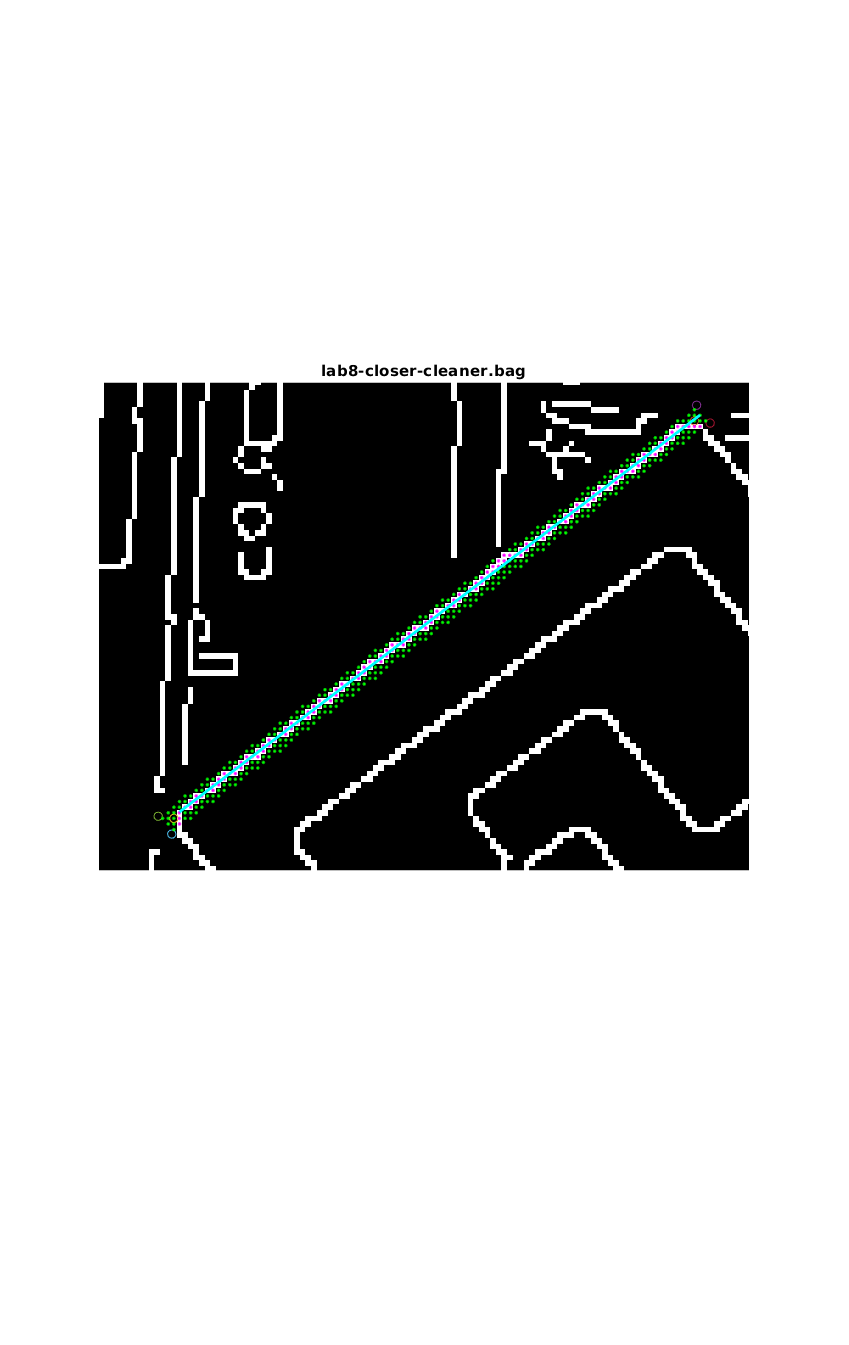}}~
\hspace{2pt}%
\subfloat[]{%
\label{fig:CannyLineFitted}%
\includegraphics[width=0.5\columnwidth, trim={2cm 1.7cm 2cm 0.7cm},clip]{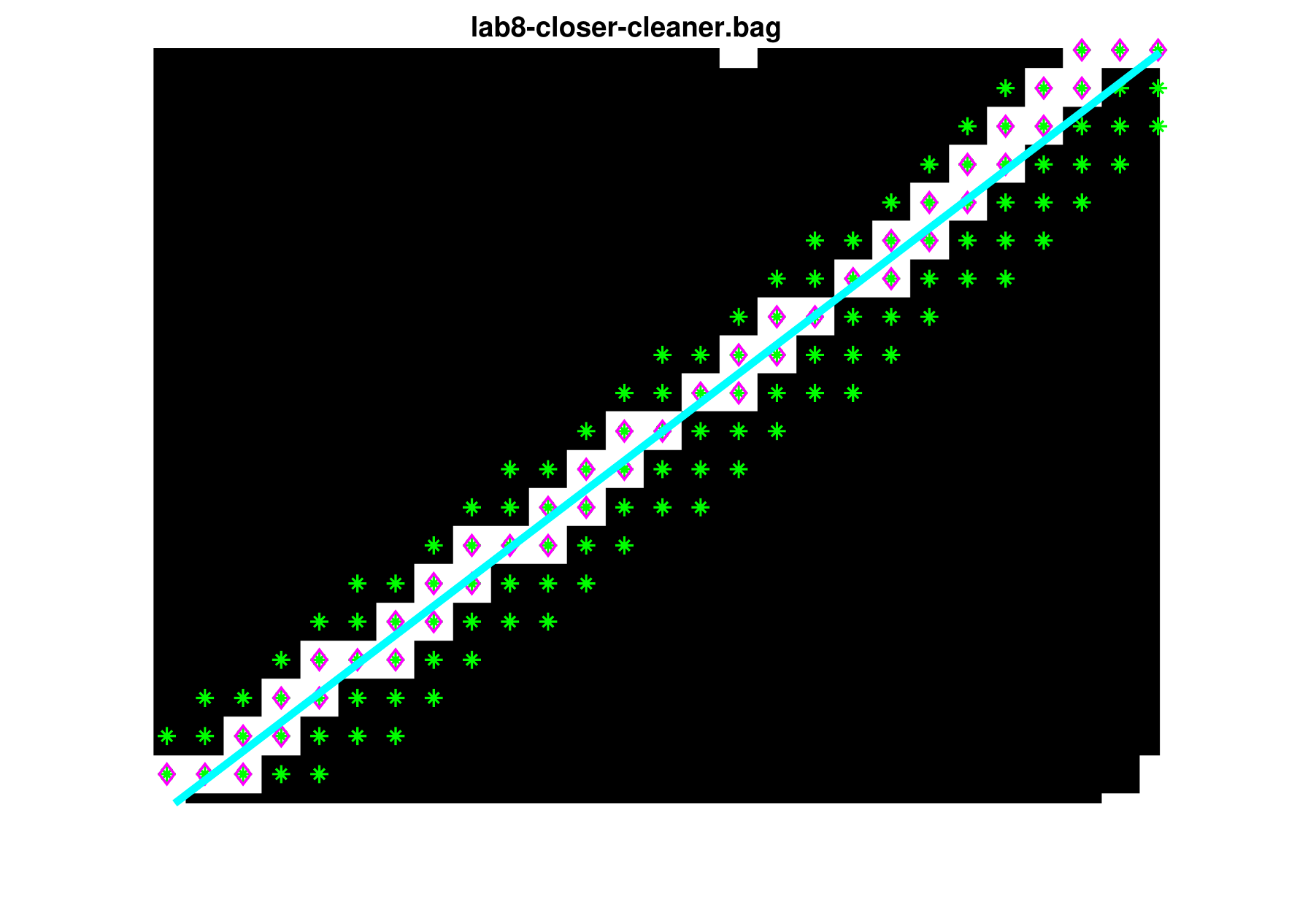}}%
\caption[]{\subref{fig:Cannyedges} shows the result of edge detection.
\subref{fig:CannyLineFitted} shows the interior pixels (marked in green) of a
bounding box given two clicked corners. The edge points and the edge line (as found by
\textit{RANSAC}) are marked in magenta and cyan, respectively. The corners are defined
by the intersections of the resulting lines.}% \label{fig:ImageEdgeDetection}%
\end{figure}

The process of determining the target corners begins by clicking near them in the
image. This is the only manual intervention required in the paper and even this
process will soon be automated. As shown in Fig.~\ref{fig:Cannyedges},
between any given two clicked points, i.e., an edge of the target, a bounding box is
drawn around the two points. Once we have roughly located the corners of the target,
we process the image with Canny edge detection\footnote{We use the \textit{edge} command in MATLAB with the `Canny' option.} to detect the edge points within the bounding box. A line is then fit through
each edge using \textit{RANSAC},  and the intersections of the resulting lines define
the target's corners, as shown in Fig.~\ref{fig:CannyLineFitted}. A video and an
implementation of this process is released along with the code.

\subsection{Extrinsic Calibration}
\label{sec:Extrinsic}
On the basis of the associated target vertices and camera-image corners, both the PnP
and IoU methods are used to find the rigid-body transformation from LiDAR to
camera. Because the
results are similar, we report only the PnP method in Table~\ref{tab:cross-v1} and then include both in the summary table (Table~\ref{tab:SummaryTable}). SVD and \textit{RANSAC} were computed using their corresponding
MATLAB commands, while the optimizations in \eqref{eq:OurNewCost}, \eqref{eq:PnP} and \eqref{eq:IoUOptimization} were done with \texttt{fmincon}.

\subsection{Computation Performance}
The calibration is done offline. The round-robin cross-validation study given in Table~\ref{tab:cross-v1} was generated in MATLAB in less than an hour. Each dataset (including the baseline and the proposed method) takes about 1.5 minutes in MATLAB.

\subsection{Quantitative Results and Round-robin Analysis}
\label{sec:QuantitativeResultS}
In Table~\ref{tab:cross-v1}, we present the RMS error of the LiDAR vertices projected
into the camera's image plane for the baseline\footnote{SVD to extract the normal and
\textit{RANSAC} for individual line fitting, with the vertices obtained as the
intersections of the lines.}, labeled RANSAC-normal (\textbf{RN}), and our method in
\eqref{eq:JKHcost} and \eqref{eq:OurNewCost}, labeled geometry-L$_1$ (\textbf{GL$_1$}). In the case of
two targets, a full round-robin study is performed: the rigid-body transformation
from LiDAR to camera is ``fit'' to the combined set of eight vertices from both
targets and then ``validated'' on the eight vertices of each of the remaining six
scenes. 

A complete round-robin study for four targets from two scenes would require 21
validations, while for six and eight targets, 35 validations each would be required.
For space reasons, we report only a subset of these possibilities. 

\begin{figure*}[b!]%
\centering
\subfloat{%
% \label{fig:edges}%
\includegraphics[width=0.69\columnwidth, trim={0.5cm 1.0cm 0cm 0.5cm},clip]{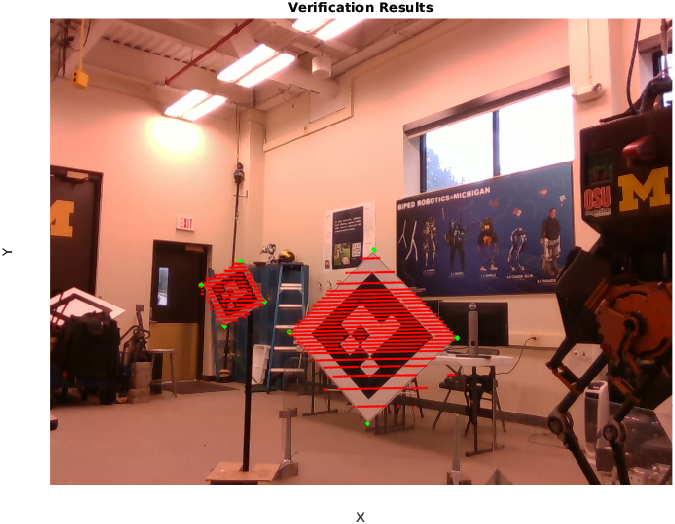}}~
\hspace{2pt}%
\subfloat{%
% \label{fig:edges}%
\includegraphics[width=0.69\columnwidth, trim={0.5cm 1.0cm 0cm 0.5cm},clip]{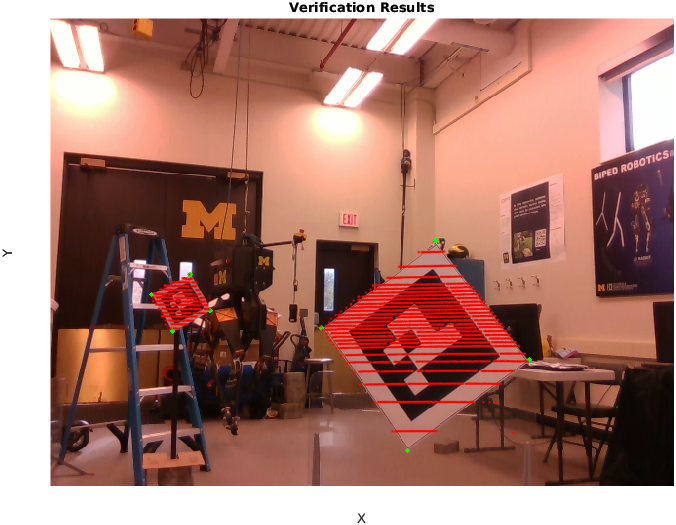}}~
\hspace{2pt}%
\subfloat{%
% \label{fig:filled}%
\includegraphics[width=0.69\columnwidth, trim={0.5cm 1.0cm 0cm 0.5cm},clip]{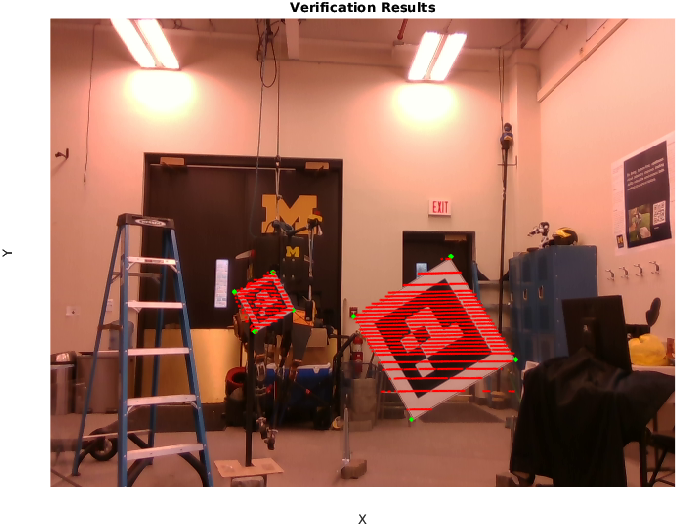}}\\
\hspace{2pt}%
\subfloat{%
% \label{fig:fiducial_edges}%
\includegraphics[width=0.69\columnwidth, trim={0.5cm 1.0cm 0cm 0.5cm},clip]{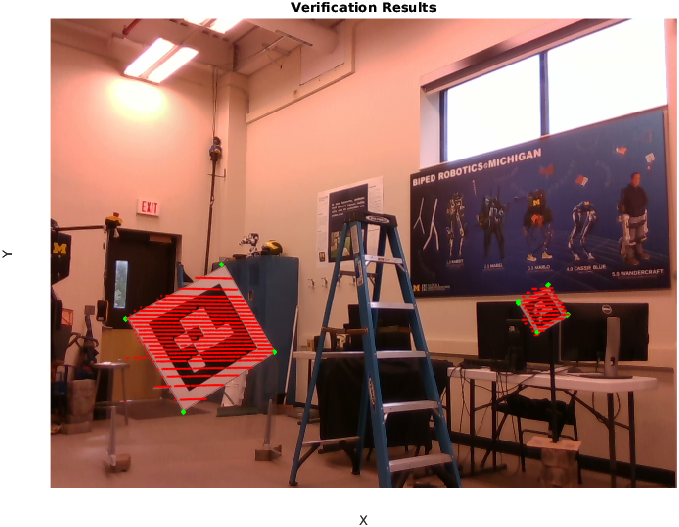}}~
\hspace{2pt}%
\subfloat{%
% \label{fig:fiducial_edges}%
\includegraphics[width=0.69\columnwidth, trim={0.5cm 1.0cm 0cm 0.5cm},clip]{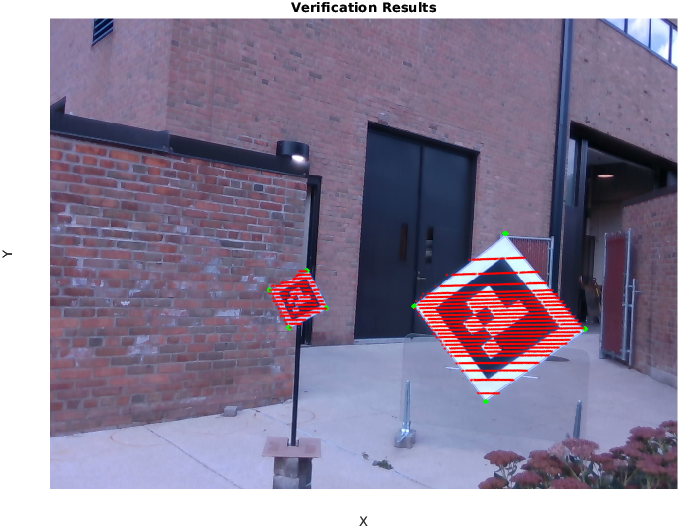}}~
% \hspace{3pt}%
% \subfloat[]{%
% \label{fig:fiducial_edges}%
% \includegraphics[width=0.5\columnwidth]{v7.png}}~
\hspace{2pt}%
\subfloat{%
% \label{fig:color_points}%
\includegraphics[width=0.68\columnwidth, trim={0.5cm 1.0cm 0cm 0.5cm},clip]{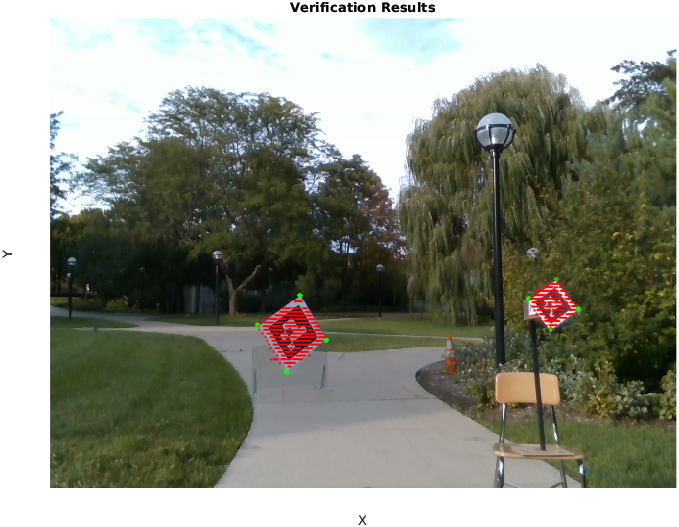}}%
\caption[]{Visual depiction of the validation data in the last row of
Table~\ref{tab:cross-v1}. For the method \textbf{GL$_1$}, five sets of estimated
LiDAR vertices for each target have been projected into the image plane and marked in
green, while the target's point cloud has been marked in red. Blowing up an image
allows the numbers reported in the table to be visualized. The vertices are key.}%
\label{fig:quantitative_results}%
\end{figure*}

To be clear, we are reporting in units of pixel per corner
\begin{equation}
\label{eq:ErrorMeasure}
\resizebox{0.4\columnwidth}{!}{%
$\sqrt{\frac{1}{4n}\sum_{i=1}^{4n}\norm{\pre[L]Y_i-\pre[C]Y_i}_2^2},$
}
\end{equation}
where $4n$ is the total number of target corners. In the case of two targets and one
scene, $n=2$. In the case of six targets from three scenes, $n=6$.
Figure~\ref{fig:quantitative_results} illustrates several point clouds projected to
the corresponding camera images. Summary data is presented in Table~\ref{tab:SummaryTable}.

\begin{table*}[]
\caption{Fitting and validation data. The gray boxes denote the fitting set and the white boxes contain validation data. The numbers are the RMS errors of the LiDAR vertices projected to the image plane, measured in units of pixels per corner; see \eqref{eq:ErrorMeasure}. For two targets, a complete round-robin study was done. For more targets, the number of combinations became prohibitive. S$_1$ through S$_7$ denote scene (experiment) number. The mean and standard deviation of each row---excluding the fitting set---are given in the last two columns.}
\label{tab:cross-v1}
\medskip
\centering
\resizebox{2\columnwidth}{!}{%
\begin{tabular}{|c|c|c|c|c|c|c|c|c|c|c|c|}
\hline
\rowcolor[HTML]{EFEFEF} 
Fitting\textbackslash{}Validation & Method & \# Tag & S1 & S2 & S3 & S4 & S5 & S6 & S7 & mean & std \\ \hline
\cellcolor[HTML]{EFEFEF} & RN & 2 & \cellcolor[HTML]{C0C0C0}{\color[HTML]{333333} 2.6618} & 8.7363 & 4.4712 & 7.3851 & 4.1269 & 7.1884 & 11.9767 & 7.3141 & 2.8991 \\ \cline{2-12} 
\multirow{-2}{*}{\cellcolor[HTML]{EFEFEF}S1} & GL$_1$ & 2 & \cellcolor[HTML]{C0C0C0}0.7728 & \textbf{2.3303} & \textbf{2.3230} & \textbf{1.6318} & \textbf{1.3694} & \textbf{1.9637} & \textbf{2.7427} & \textbf{2.0602} & \textbf{0.5055} \\ \hline
\cellcolor[HTML]{EFEFEF} & RN & 2 & 3.3213 & \cellcolor[HTML]{C0C0C0}7.6645 & 4.9169 & 5.2951 & 4.0811 & 4.4345 & 7.7397 & 4.9648 & 1.5215 \\ \cline{2-12} 
\multirow{-2}{*}{\cellcolor[HTML]{EFEFEF}S2} & GL$_1$ & 2 & \textbf{2.1368} & \cellcolor[HTML]{C0C0C0}1.5181 & \textbf{3.2027} & \textbf{3.2589} & \textbf{2.4480} & \textbf{4.1563} & \textbf{4.4254} & \textbf{3.2713} & \textbf{0.9039} \\ \hline
\cellcolor[HTML]{EFEFEF} & RN & 2 & 4.9909 & 9.5620 & \cellcolor[HTML]{C0C0C0}3.6271 & 8.7533 & 4.6421 & 10.1308 & 15.7764 & 8.9759 & 4.0654 \\ \cline{2-12} 
\multirow{-2}{*}{\cellcolor[HTML]{EFEFEF}S3} & GL$_1$ & 2 & \textbf{4.6720} & \textbf{5.9350} & \cellcolor[HTML]{C0C0C0}1.8469 & \textbf{6.9331} & \textbf{4.4352} & \textbf{8.9493} & \textbf{15.3862} & \textbf{7.7185} & \textbf{4.1029} \\ \hline
\cellcolor[HTML]{EFEFEF} & RN & 2 & 21.2271 & 22.1641 & 17.5779 & \cellcolor[HTML]{C0C0C0}2.9452 & 15.9909 & 8.8797 & 15.3636 & 16.8672 & 4.7833 \\ \cline{2-12} 
\multirow{-2}{*}{\cellcolor[HTML]{EFEFEF}S4} & GL$_1$ & 2 & \textbf{4.3681} & \textbf{4.7176} & \textbf{4.4804} & \cellcolor[HTML]{C0C0C0}0.4986 & \textbf{3.9004} & \textbf{3.3891} & \textbf{3.7440} & \textbf{4.0999} & \textbf{0.5040} \\ \hline
\cellcolor[HTML]{EFEFEF} & RN & 2 & 3.4621 & 8.3131 & 4.8500 & 7.6217 & \cellcolor[HTML]{C0C0C0}3.5197 & 7.4838 & 12.4364 & 7.3612 & 3.1066 \\ \cline{2-12} 
\multirow{-2}{*}{\cellcolor[HTML]{EFEFEF}S5} & GL$_1$ & 2 & \textbf{2.0590} & \textbf{2.9541} & \textbf{3.0224} & \textbf{3.5483} & \cellcolor[HTML]{C0C0C0}0.7392 & \textbf{3.1386} & \textbf{6.0885} & \textbf{3.4685} & \textbf{1.3733} \\ \hline
\cellcolor[HTML]{EFEFEF} & RN & 2 & 29.4400 & 27.5404 & 27.9955 & 9.7005 & 20.6511 & \cellcolor[HTML]{C0C0C0}1.4253 & 9.5050 & 20.8054 & 9.1941 \\ \cline{2-12} 
\multirow{-2}{*}{\cellcolor[HTML]{EFEFEF}S6} & GL$_1$ & 2 & \textbf{5.1207} & \textbf{5.0574} & \textbf{5.3537} & \textbf{2.0539} & \textbf{4.1739} & \cellcolor[HTML]{C0C0C0}1.0012 & \textbf{2.4194} & \textbf{4.0298} & \textbf{1.4503} \\ \hline
\cellcolor[HTML]{EFEFEF} & RN & 2 & 7.7991 & 9.9647 & 7.6857 & 4.1640 & 6.1619 & 2.3398 & \cellcolor[HTML]{C0C0C0}2.3708 & 6.3525 & 2.7512 \\ \cline{2-12} 
\multirow{-2}{*}{\cellcolor[HTML]{EFEFEF}S7} & GL$_1$ & 2 & \textbf{2.1563} & \textbf{2.7837} & \textbf{3.1058} & \textbf{1.3234} & \textbf{2.4537} & \textbf{2.0838} & \cellcolor[HTML]{C0C0C0}1.5389 & \textbf{2.3178} & \textbf{0.6207} \\ \hhline{|=|=|=|=|=|=|=|=|=|=|=|=|}
\cellcolor[HTML]{EFEFEF} & RN & 4 & 6.9426 & 9.6281 & 6.9136 & 3.9650 & 5.6420 & \cellcolor[HTML]{C0C0C0}2.2399 & \cellcolor[HTML]{C0C0C0}2.2399 & 6.6183 & 2.0764 \\ \cline{2-12} 
\multirow{-2}{*}{\cellcolor[HTML]{EFEFEF}S6-S7} & GL$_1$ & 4 & \textbf{2.2409} & \textbf{2.5607} & \textbf{2.9773} & \textbf{1.8215} & \textbf{2.1995} & \cellcolor[HTML]{C0C0C0}0.3417 & \cellcolor[HTML]{C0C0C0}0.3417 & \textbf{2.3600} & \textbf{0.4334} \\ \hline
\cellcolor[HTML]{EFEFEF} & RN & 4 & 4.2476 & 8.5054 & 4.6712 & \cellcolor[HTML]{C0C0C0}2.8686 & 4.3311 & \cellcolor[HTML]{C0C0C0}{\color[HTML]{333333} 2.8686} & 2.8748 & 4.9260 & 2.1153 \\ \cline{2-12} 
\multirow{-2}{*}{\cellcolor[HTML]{EFEFEF}S4-S6} & GL$_1$ & 4 & \textbf{1.0746} & \textbf{1.8871} & \textbf{2.3619} & \cellcolor[HTML]{C0C0C0}0.2341 & \textbf{1.5350} & \cellcolor[HTML]{C0C0C0}{\color[HTML]{333333} 0.2341} & \textbf{2.6191} & \textbf{1.8955} & \textbf{0.6215} \\ \hline
\cellcolor[HTML]{EFEFEF} & RN & 4 & 2.9309 & 8.0801 & 4.1196 & 3.4985 & \cellcolor[HTML]{C0C0C0}3.1519 & \cellcolor[HTML]{C0C0C0}3.1519 & 3.2204 & 4.3699 & 2.1201 \\ \cline{2-12} 
\multirow{-2}{*}{\cellcolor[HTML]{EFEFEF}S5-S6} & GL$_1$ & 4 & \textbf{1.1541} & \textbf{2.1026} & \textbf{2.5017} & \textbf{1.4267} & \cellcolor[HTML]{C0C0C0}0.3291 & \cellcolor[HTML]{C0C0C0}0.3291 & \textbf{2.5693} & \textbf{1.9509} & \textbf{0.6361} \\ \hline
\cellcolor[HTML]{EFEFEF} & RN & 4 & 3.1991 & \cellcolor[HTML]{C0C0C0}{\color[HTML]{333333} 6.0812} & 4.7311 & 5.4558 & \cellcolor[HTML]{C0C0C0}{\color[HTML]{333333} 6.0812} & 4.8344 & 8.6576 & 5.3756 & 2.0139 \\ \cline{2-12} 
\multirow{-2}{*}{\cellcolor[HTML]{EFEFEF}S2-S5} & GL$_1$ & 4 & \textbf{0.9538} & \cellcolor[HTML]{C0C0C0}{\color[HTML]{333333} 0.4803} & \textbf{2.3810} & \textbf{1.4556} & \cellcolor[HTML]{C0C0C0}{\color[HTML]{333333} 0.4803} & \textbf{1.1592} & \textbf{2.5277} & \textbf{1.6955} & \textbf{0.7172} \\ \hline
\cellcolor[HTML]{EFEFEF} & RN & 4 & 2.8820 & \cellcolor[HTML]{C0C0C0}6.2167 & \cellcolor[HTML]{C0C0C0}6.2167 & 4.4672 & 3.9536 & 3.7802 & 6.2324 & 4.2631 & 1.2406 \\ \cline{2-12} 
\multirow{-2}{*}{\cellcolor[HTML]{EFEFEF}S2-S3} & GL$_1$ & 4 & \textbf{1.0896} & \cellcolor[HTML]{C0C0C0}0.4826 & \cellcolor[HTML]{C0C0C0}0.4826 & \textbf{1.5349} & \textbf{1.5938} & \textbf{1.5597} & \textbf{2.7543} & \textbf{1.7065} & \textbf{0.6209} \\ \hline
\cellcolor[HTML]{EFEFEF} & RN & 4 & \cellcolor[HTML]{C0C0C0}2.8595 & 8.2471 & 4.3297 & 3.7712 & 4.1023 & 2.4679 & \cellcolor[HTML]{C0C0C0}2.8595 & 4.5836 & 2.1710 \\ \cline{2-12} 
\multirow{-2}{*}{\cellcolor[HTML]{EFEFEF}S1-S7} & GL$_1$ & 4 & \cellcolor[HTML]{C0C0C0}1.3542 & \textbf{1.9973} & \textbf{2.3987} & \textbf{1.5926} & \textbf{1.4978} & \textbf{1.6130} & \cellcolor[HTML]{C0C0C0}1.3542 & \textbf{1.8199} & \textbf{0.3757} \\ \hline
\cellcolor[HTML]{EFEFEF} & RN & 4 & \cellcolor[HTML]{C0C0C0}2.5633 & 8.0573 & 4.1762 & 3.6859 & 3.8973 & \cellcolor[HTML]{C0C0C0}2.5633 & 3.2569 & 4.6147 & 1.9535 \\ \cline{2-12} 
\multirow{-2}{*}{\cellcolor[HTML]{EFEFEF}S1-S6} & GL$_1$ & 4 & \cellcolor[HTML]{C0C0C0}0.9753 & \textbf{1.9154} & \textbf{2.3361} & \textbf{1.2906} & \textbf{1.2827} & \cellcolor[HTML]{C0C0C0}0.9753 & \textbf{2.3205} & \textbf{1.8291} & \textbf{0.5231} \\ \hhline{|=|=|=|=|=|=|=|=|=|=|=|=|}
\cellcolor[HTML]{EFEFEF} & RN & 6 & 2.9824 & 8.0943 & 4.1273 & \cellcolor[HTML]{C0C0C0}3.2821 & \cellcolor[HTML]{C0C0C0}3.2821 & \cellcolor[HTML]{C0C0C0}3.2821 & 3.0841 & 4.5720 & 2.4045 \\ \cline{2-12} 
\multirow{-2}{*}{\cellcolor[HTML]{EFEFEF}S4-S5-S6} & GL$_1$ & 6 & \textbf{0.9713} & \textbf{1.9887} & \textbf{2.4395} & \cellcolor[HTML]{C0C0C0}0.3288 & \cellcolor[HTML]{C0C0C0}0.3288 & \cellcolor[HTML]{C0C0C0}0.3288 & \textbf{2.4369} & \textbf{1.9591} & \textbf{0.6918} \\ \hline
\cellcolor[HTML]{EFEFEF} & RN & 6 & 2.7816 & \cellcolor[HTML]{C0C0C0}5.5382 & \cellcolor[HTML]{C0C0C0}5.5382 & 4.5817 & \cellcolor[HTML]{C0C0C0}5.5382 & 3.8262 & 6.6229 & 4.4531 & 1.6239 \\ \cline{2-12} 
\multirow{-2}{*}{\cellcolor[HTML]{EFEFEF}S2-S3-S5} & GL$_1$ & 6 & \textbf{1.0022} & \cellcolor[HTML]{C0C0C0}0.4712 & \cellcolor[HTML]{C0C0C0}0.4712 & \textbf{1.4755} & \cellcolor[HTML]{C0C0C0}0.4712 & \textbf{1.1722} & \textbf{2.7791} & \textbf{1.6073} & \textbf{0.8054} \\ \hline
\cellcolor[HTML]{EFEFEF} & RN & 6 & \cellcolor[HTML]{C0C0C0}3.3906 & 8.1677 & \cellcolor[HTML]{C0C0C0}3.3906 & 3.6688 & 3.9558 & 2.2913 & \cellcolor[HTML]{C0C0C0}3.3906 & 4.5209 & 2.5374 \\ \cline{2-12} 
\multirow{-2}{*}{\cellcolor[HTML]{EFEFEF}S1-S3-S7} & GL$_1$ & 6 & \cellcolor[HTML]{C0C0C0}1.7666 & \textbf{1.9994} & \cellcolor[HTML]{C0C0C0}1.7666 & \textbf{1.5854} & \textbf{1.4754} & \textbf{1.5124} & \cellcolor[HTML]{C0C0C0}1.7666 & \textbf{1.6432} & \textbf{0.2419} \\ \hline
\cellcolor[HTML]{EFEFEF} & RN & 6 & 2.9428 & \cellcolor[HTML]{C0C0C0}5.5357 & \cellcolor[HTML]{C0C0C0}5.5357 & \cellcolor[HTML]{C0C0C0}5.5357 & 3.8913 & 2.4828 & 3.7673 & 3.2711 & 0.6733 \\ \cline{2-12} 
\multirow{-2}{*}{\cellcolor[HTML]{EFEFEF}S2-S3-S4} & GL$_1$ & 6 & \textbf{0.9523} & \cellcolor[HTML]{C0C0C0}1.8195 & \cellcolor[HTML]{C0C0C0}1.8195 & \cellcolor[HTML]{C0C0C0}1.8195 & \textbf{1.5275} & \textbf{1.3631} & \textbf{2.4347} & \textbf{1.5694} & \textbf{0.6255} \\ \hline
\cellcolor[HTML]{EFEFEF} & RN & 6 & \cellcolor[HTML]{C0C0C0}5.3281 & \cellcolor[HTML]{C0C0C0}5.3281 & \cellcolor[HTML]{C0C0C0}5.3281 & 4.5086 & 3.8577 & 3.7165 & 6.4413 & 4.6310 & 1.2552 \\ \cline{2-12} 
\multirow{-2}{*}{\cellcolor[HTML]{EFEFEF}S1-S2-S3} & GL$_1$ & 6 & \cellcolor[HTML]{C0C0C0}1.7755 & \cellcolor[HTML]{C0C0C0}1.7755 & \cellcolor[HTML]{C0C0C0}1.7755 & \textbf{1.3698} & \textbf{1.4807} & \textbf{1.3422} & \textbf{2.3156} & \textbf{1.6271} & \textbf{0.4629} \\ \hline
\cellcolor[HTML]{EFEFEF} & RN & 6 & 3.2743 & 8.2067 & 4.2628 & 3.5584 & \cellcolor[HTML]{C0C0C0}3.0628 & \cellcolor[HTML]{C0C0C0}3.0628 & \cellcolor[HTML]{C0C0C0}3.0628 & 4.8256 & 2.2921 \\ \cline{2-12} 
\multirow{-2}{*}{\cellcolor[HTML]{EFEFEF}S5-S6-S7} & GL$_1$ & 6 & \textbf{0.9594} & \textbf{2.0149} & \textbf{2.3877} & \textbf{1.3754} & \cellcolor[HTML]{C0C0C0}1.4905 & \cellcolor[HTML]{C0C0C0}1.4905 & \cellcolor[HTML]{C0C0C0}1.4905 & \textbf{1.6843} & \textbf{0.6390} \\ \hhline{|=|=|=|=|=|=|=|=|=|=|=|=|}
\cellcolor[HTML]{EFEFEF} & RN & 8 & \cellcolor[HTML]{C0C0C0}3.5316 & 8.1362 & \cellcolor[HTML]{C0C0C0}3.5316 & 3.4616 & \cellcolor[HTML]{C0C0C0}3.5316 & 2.2642 & \cellcolor[HTML]{C0C0C0}3.5316 & 4.6207 & 3.1029 \\ \cline{2-12} 
\multirow{-2}{*}{\cellcolor[HTML]{EFEFEF}S1-S3-S5-S7} & GL$_1$ & 8 & \cellcolor[HTML]{C0C0C0}1.6904 & \textbf{1.9886} & \cellcolor[HTML]{C0C0C0}1.6904 & \textbf{1.4743} & \cellcolor[HTML]{C0C0C0}1.6904 & \textbf{1.4136} & \cellcolor[HTML]{C0C0C0}1.6904 & \textbf{1.6255} & \textbf{0.3159} \\ \hline
\cellcolor[HTML]{EFEFEF} & RN & 8 & \cellcolor[HTML]{C0C0C0}4.9240 & \cellcolor[HTML]{C0C0C0}4.9240 & 4.2049 & 3.6159 & \cellcolor[HTML]{C0C0C0}4.9240 & 2.2928 & \cellcolor[HTML]{C0C0C0}4.9240 & 3.3712 & 0.9793 \\ \cline{2-12} 
\multirow{-2}{*}{\cellcolor[HTML]{EFEFEF}S1-S2-S5-S7} & GL$_1$ & 8 & \cellcolor[HTML]{C0C0C0}1.5269 & \cellcolor[HTML]{C0C0C0}1.5269 & \textbf{2.4070} & \textbf{1.4788} & \cellcolor[HTML]{C0C0C0}1.5269 & \textbf{1.4650} & \cellcolor[HTML]{C0C0C0}1.5269 & \textbf{1.7836} & \textbf{0.5399} \\ \hline
\cellcolor[HTML]{EFEFEF} & RN & 8 & \cellcolor[HTML]{C0C0C0}3.6113 & 8.0593 & \cellcolor[HTML]{C0C0C0}3.6113 & \cellcolor[HTML]{C0C0C0}3.6113 & \cellcolor[HTML]{C0C0C0}3.6113 & 2.4532 & 3.8397 & 4.7841 & 2.9199 \\ \cline{2-12} 
\multirow{-2}{*}{\cellcolor[HTML]{EFEFEF}S1-S3-S4-S5} & GL$_1$ & 8 & \cellcolor[HTML]{C0C0C0}1.4986 & \textbf{2.0563} & \cellcolor[HTML]{C0C0C0}1.4986 & \cellcolor[HTML]{C0C0C0}1.4986 & \cellcolor[HTML]{C0C0C0}1.4986 & \textbf{1.2256} & \textbf{2.6797} & \textbf{1.9872} & \textbf{0.7295} \\ \hline
\cellcolor[HTML]{EFEFEF} & RN & 8 & 3.0803 & \cellcolor[HTML]{C0C0C0}5.0057 & 4.1893 & \cellcolor[HTML]{C0C0C0}5.0057 & \cellcolor[HTML]{C0C0C0}5.0057 & 2.3307 & \cellcolor[HTML]{C0C0C0}5.0057 & 3.2001 & 0.9351 \\ \cline{2-12} 
\multirow{-2}{*}{\cellcolor[HTML]{EFEFEF}S2-S4-S5-S7} & GL$_1$ & 8 & \textbf{0.9530} & \cellcolor[HTML]{C0C0C0}1.5921 & \textbf{2.4280} & \cellcolor[HTML]{C0C0C0}1.5921 & \cellcolor[HTML]{C0C0C0}1.5921 & \textbf{1.4865} & \cellcolor[HTML]{C0C0C0}1.5921 & \textbf{1.6225} & \textbf{0.7469} \\ \hline
\end{tabular}
}
\end{table*}

% \begin{table}[]
% \caption{A summary of validation data in Table \ref{tab:cross-v1}. This table compares mean and standard deviation for baseline and our approach as a function of the number of targets used in training. Units are pixel per corner.}
% \label{tab:SummaryTable}
% \centering
% \resizebox{1\columnwidth}{!}{%
% \begin{tabular}{|c|c|c|c|c|c|}
% \hline
% \rowcolor[HTML]{EFEFEF} 
%  & \# Tag & 2 & 4 & 6 & 8 \\ \hline
% \cellcolor[HTML]{EFEFEF}RN & mean & 10.3773 & 4.9645 & 4.3789 & 3.9940 \\ \hline
% \cellcolor[HTML]{EFEFEF}GL$_1$ & mean & \textbf{3.8523} & \textbf{1.8939} & \textbf{1.6817} & \textbf{1.7547} \\ \hline
% \cellcolor[HTML]{EFEFEF}RN & std & 7.0887 & 1.9532 & 1.7771 & 2.0467 \\ \hline
% \cellcolor[HTML]{EFEFEF}GL$_1$ & std & \textbf{2.4155} & \textbf{0.5609} & \textbf{0.5516} & \textbf{0.5419} \\ \hline
% \end{tabular}
% }
% \end{table}

\begin{table}[]
\caption{A summary of validation data in Table \ref{tab:cross-v1}. This table compares mean and standard deviation for baseline and our approach as a function of the number of targets used in fitting. Units are pixel per corner.}
\label{tab:SummaryTable}
\centering
\resizebox{1\columnwidth}{!}{%
\begin{tabular}{|
>{\columncolor[HTML]{EFEFEF}}c |c|l|l|l|l|}
\hline
Method    & \cellcolor[HTML]{EFEFEF}\# Tag & \multicolumn{1}{c|}{\cellcolor[HTML]{EFEFEF}2}                 & \multicolumn{1}{c|}{\cellcolor[HTML]{EFEFEF}4}                 & \multicolumn{1}{c|}{\cellcolor[HTML]{EFEFEF}6}                 & \multicolumn{1}{c|}{\cellcolor[HTML]{EFEFEF}8}                 \\ \hline
Baseline-RN  & mean                           & \cellcolor[HTML]{FFFFFF}{\color[HTML]{24292E} 10.3773}         & \cellcolor[HTML]{FFFFFF}{\color[HTML]{24292E} 4.9645}          & \cellcolor[HTML]{FFFFFF}{\color[HTML]{24292E} 4.3789}          & \cellcolor[HTML]{FFFFFF}{\color[HTML]{24292E} 3.9940}          \\ \hline
GL$_1$-PnP & mean                           & \cellcolor[HTML]{F6F8FA}{\color[HTML]{24292E} \textbf{3.8523}} & \cellcolor[HTML]{F6F8FA}{\color[HTML]{24292E} \textbf{1.8939}} & \cellcolor[HTML]{F6F8FA}{\color[HTML]{24292E} \textbf{1.6817}} & \cellcolor[HTML]{F6F8FA}{\color[HTML]{24292E} \textbf{1.7547}} \\ \hline
GL$_1$-IoU & mean                           & \cellcolor[HTML]{FFFFFF}{\color[HTML]{24292E} 4.9019}          & \cellcolor[HTML]{FFFFFF}{\color[HTML]{24292E} 2.2442}          & \cellcolor[HTML]{FFFFFF}{\color[HTML]{24292E} 1.7631}          & \cellcolor[HTML]{FFFFFF}{\color[HTML]{24292E} 1.7837}          \\ \hline
Baseline-RN  & std                            & \cellcolor[HTML]{F6F8FA}{\color[HTML]{24292E} 7.0887}          & \cellcolor[HTML]{F6F8FA}{\color[HTML]{24292E} 1.9532}          & \cellcolor[HTML]{F6F8FA}{\color[HTML]{24292E} 1.7771}          & \cellcolor[HTML]{F6F8FA}{\color[HTML]{24292E} 2.0467}          \\ \hline
GL$_1$-PnP & std                            & \cellcolor[HTML]{FFFFFF}{\color[HTML]{24292E} \textbf{2.4155}} & \cellcolor[HTML]{FFFFFF}{\color[HTML]{24292E} \textbf{0.5609}} & \cellcolor[HTML]{FFFFFF}{\color[HTML]{24292E} 0.5516}          & \cellcolor[HTML]{FFFFFF}{\color[HTML]{24292E} 0.5419}          \\ \hline
GL$_1$-IoU & std                            & \cellcolor[HTML]{F6F8FA}{\color[HTML]{24292E} 2.5060}          & \cellcolor[HTML]{F6F8FA}{\color[HTML]{24292E} 0.7162}          & \cellcolor[HTML]{F6F8FA}{\color[HTML]{24292E} \textbf{0.5070}} & \cellcolor[HTML]{F6F8FA}{\color[HTML]{24292E} \textbf{0.4566}} \\ \hline
\end{tabular}
}
\end{table}

\section{Qualitative Results and Discussion} 
\label{sec:QualitativeResults}

\begin{figure*}[t!]%
\centering
\subfloat{%
% \label{fig:edges}%
\includegraphics[width=0.66\columnwidth, trim={0.5cm 1.0cm 0cm 0.5cm},clip]{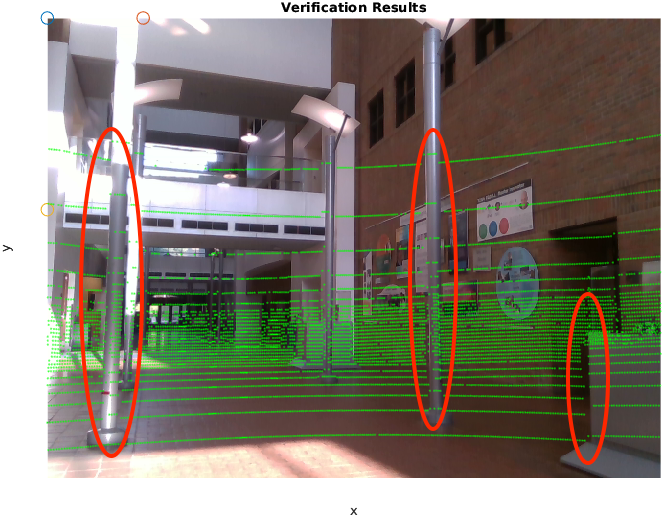}}~
\hspace{3pt}%
\subfloat{%
% \label{fig:edges}%
\includegraphics[width=0.66\columnwidth, trim={0.5cm 1.0cm 0cm 0.5cm},clip]{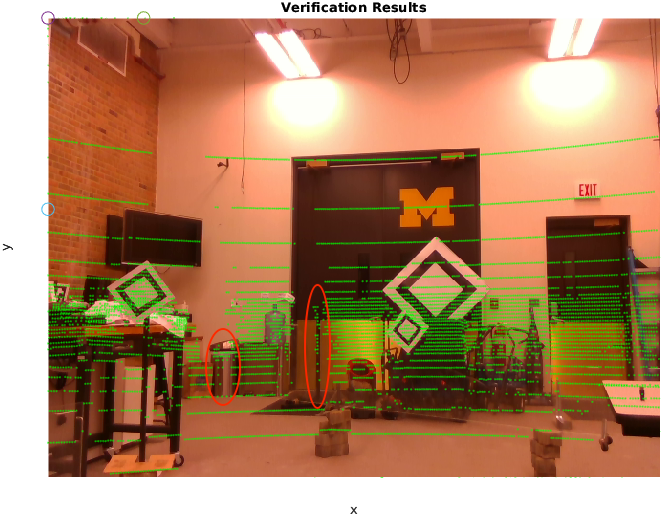}}~
\hspace{3pt}%
\subfloat{%
% \label{fig:edges}%
\includegraphics[width=0.67\columnwidth, trim={0.5cm 1.0cm 0cm 0.5cm},clip]{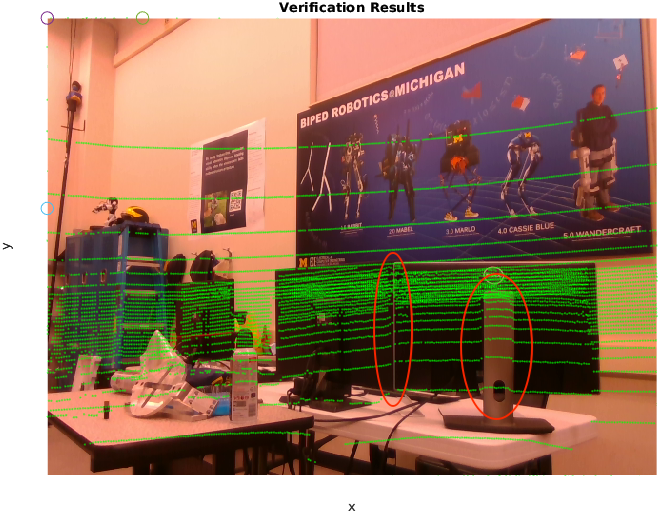}}\\
% \hspace{3pt}%
% \subfloat[]{%
% \label{fig:filled}%
% \includegraphics[width=0.5\columnwidth]{test4.png}}
\hspace{3pt}%
\subfloat{%
% \label{fig:fiducial_edges}%
\includegraphics[width=0.67\columnwidth, trim={0.5cm 1.0cm 0cm 0.5cm},clip]{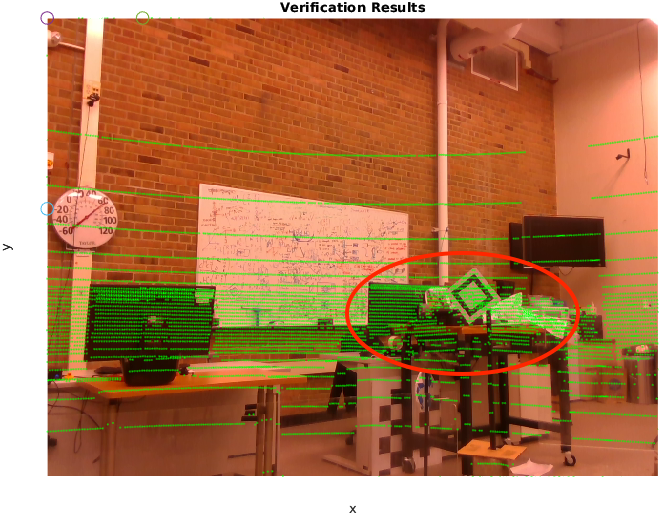}}~
\hspace{3pt}%
\subfloat{%
% \label{fig:fiducial_edges}%
\includegraphics[width=0.66\columnwidth, trim={0.5cm 1.0cm 0cm 0.5cm},clip]{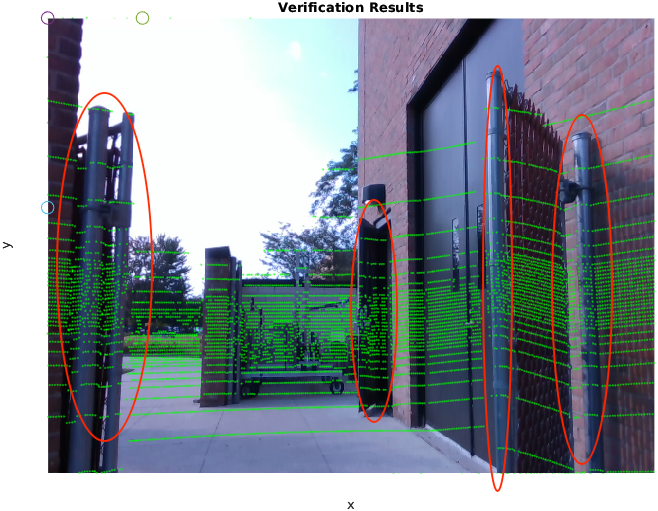}}~
% \hspace{3pt}%
% \subfloat[]{%
% \label{fig:fiducial_edges}%
% \includegraphics[width=0.5\columnwidth]{test7.png}}~
\hspace{3pt}%
\subfloat{%
% \label{fig:color_points}%
\includegraphics[width=0.66\columnwidth, trim={0.5cm 1.0cm 0cm 0.5cm},clip]{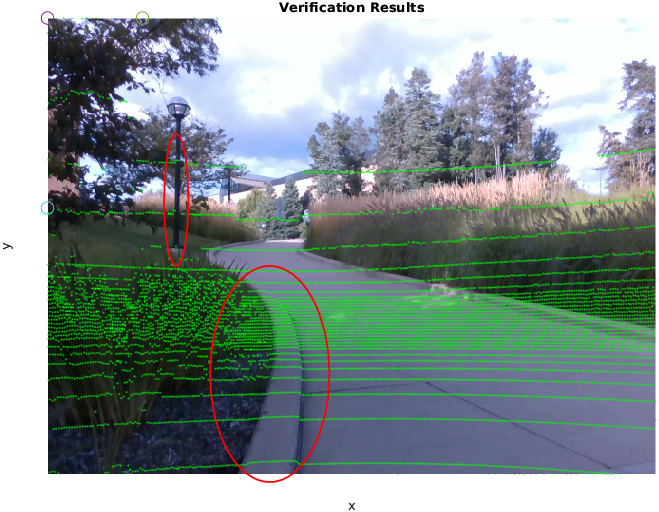}}%
\caption[]{Qualitative validation results. Using the extrinsic transformation obtained by the method \textbf{GL$_1$} applied to S$_1$, the LiDAR point cloud has been projected into the image plane for the other data sets and marked in green. The red circles highlight various poles, door edges, desk legs, monitors, and sidewalk curbs where the quality of the alignment can be best judged. The reader may find other areas of interest. Enlarge in your browser for best viewing. }% \label{fig:detector}%
\label{fig:TestingDataSet}
\end{figure*}

In LiDAR to camera calibration, due to the lack of ground truth, it is common to
show projections of LiDAR point clouds onto camera images. Often it is unclear if one
is viewing fitting data or validation data. In Figure~\ref{fig:TestingDataSet}, we
show a set of projections of LiDAR point clouds onto camera images for validation
data. An additional set of images can be found in~\cite{githubFile}. All of them show
that the key geometric features in the image and point cloud are well aligned. The
quality of the alignment has allowed our laboratory to build a high-quality
(real-time) 3D semantic map with our bipedal robot Cassie Blue \cite{gan2019bayesian}. The
map fuses LiDAR, camera, and IMU data;  with the addition of a simple planner, it led
to autonomous navigation \cite{CassieSemanticMappingYT}.

Tables~\ref{tab:cross-v1} and~\ref{tab:SummaryTable} show that \textbf{GL$_1$}
outperforms the baseline: on average, there is more than a 50\% reduction in projection error and a 70\% reduction in its variance. As for the sources of this improvement, we highlight the following points:
\begin{enumerate}
\renewcommand{\labelenumi}{(\alph{enumi})}
\setlength{\itemsep}{.2cm}
\item  Least-squares-based methods estimate a normal vector for the target's point
    cloud. As shown in Fig.~\ref{fig:FigureLiDARScanExample} and
    Fig.~\ref{fig:FigureLiDARScanExample3Rings}, even though the target is flat and
    has a well-defined normal, the returns in the LiDAR point cloud do not lie on a
    plane. Hence, a global normal vector does not exist. 

\item  Least-squares-based methods extract the target edge points from the point
    cloud for use in line fitting. The line fitting must be done in the plane defined
    by the estimated normal because, in 3D, non-parallel lines do not necessarily
    intersect to define a vertex.  

\item \textbf{GL$_1$} explicitly uses the target geometry in formulating the cost
    function. By assigning zero cost to interior points in the ``ideal target
    volume'' and non-zero otherwise, the optimization problem is focused on orienting
    the boundary of the target volume within the 3D point cloud. This perspective
    seems not to have been used before. 

\item Hence, our approach does not require the (tedious and error-prone)
\textit{explicit} extraction and assignment of points to target edges. The
determination of \textit{boundary points} in 3D is \textit{implicitly} being done
with the cost function. \end{enumerate}

At the present time, our extrinsic calibration method is not ``automatic''. The one manual intervention, namely clicking on the approximate target corners in the camera image, will be automated soon.  

We have not conducted a study on how to place the targets. This is an interesting piece of future work because of the nonlinear operation required when projecting LiDAR points to the camera plane; see \eqref{eq:projection_linear} and \eqref{eq:projection_nonlinear}.

% \begin{figure*}[b!]
%  \centering
%   %trim={<left> <lower> <right> <upper>}
%  \includegraphics[scale=0.4,trim={10.1cm 20cm 1cm 5.2cm},clip]{TargetDistribution}
%  \caption{\jkh{The exp numbers have to be changed!}}
% \end{figure*}

\section{Conclusions}
\label{sec:Conclusions}
% \bh{Do we need to mention IoU optimization in the conclusion? We spent much time and paragraph describing it. We also provide summary results but we do not discuss the results. What do you think?}
We proposed a new method to determine the extrinsic calibration of a LiDAR camera pair. When evaluated against a state-of-the-art baseline, it resulted in, on average, more than a 50\% reduction in LiDAR-to-camera projection error and a 70\% reduction in its variance. These results were established through a round-robin validation study when two targets are used in fitting, and further buttressed with results for fitting on four, six, and eight targets. 

Two other benefits of our $L_1$-based method are: (1) it does not require the estimation of a target normal vector from an inherently noisy point cloud; and (2) it also obviates the identification of edge points and their association with specific sides of a target. In combination with lower RMS error and variance, we believe our results may provide an attractive alternative to current target-based methods for extrinsic calibration.

% We proposed improvements to target-based LiDAR-camera calibration that resulted in, on average,  more than a 50\% reduction in projection error from LiDAR to camera and a 70\% reduction in its variance.We evaluated our proposed method and compared it with a state-of-the-art baseline method in a round-robin validation study when two targets are used in training and presented further data for four, six, and eight targets being used in training. In the validation study, our $L_1$-based method achieved a per corner RMS error, measured in the image plane, on the order of a few pixels, when comparing the projected LiDAR vertices to the image corners. The validation results show our method is repeatable and outperforms the baseline. 

%%%%%%%%%%%%%%%%%%%%%%%%%%%
% \balance
% \newpage
% \clearpage

% \bibliographystyle{bib/spbasic}
% \bibliographystyle{bib/IEEEtran}
% \bibliography{bib/strings-abrv,bib/ieee-abrv,bib/references}
% \bibliography{bib/strings-abrv,bib/ieee-abrv,bib/references}
%\bibliography{bib/bibJWG.bib}
\bibliographystyle{bib/IEEEtran}

\bibliography{bib/bibL2C.bib,bib/bibJWG.bib}
% \balance
\begin{IEEEbiography}[{\includegraphics[width=1in,height=1.25in,clip,keepaspectratio]{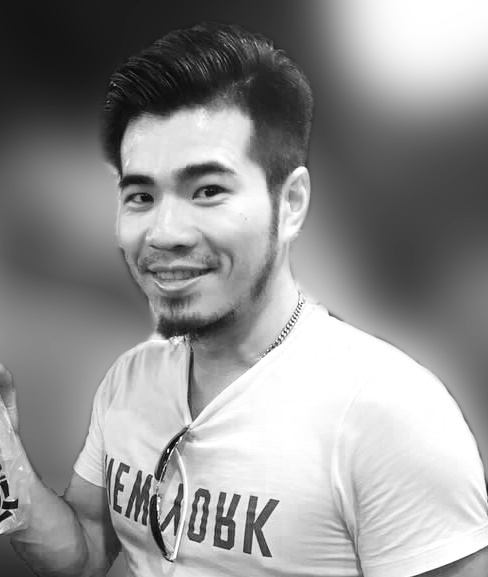}}]{Jiunn-Kai Huang} received the Master's degree in robotics from the University of Michigan, Ann Arbor in 2018.  He is currently a Ph.D. candidate in robotics at the University of Michigan, Ann Arbor. His current research interests are perception, sensor fusion and motion planning. He has also researched perception for action recognition and 3D skeleton pose estimation to assist autonomous vehicles in navigating near humans.
\end{IEEEbiography}

\begin{IEEEbiography}[{\includegraphics[width=1in,height=1.25in,clip,keepaspectratio]{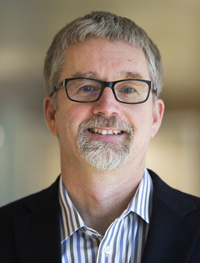}}]{Jessy W. Grizzle} (S’78–M’83–F’97) received the Ph.D. in electrical engineering from The University of Texas at Austin in 1983. He is the Director of the Michigan Robotics Institute and a Professor of Electrical Engineering and Computer Science at the University of Michigan, where he holds the titles of the Elmer Gilbert Distinguished University Professor and the Jerry and Carol Levin Professor of Engineering. He jointly holds sixteen patents dealing with emissions reduction in passenger vehicles through improved control system design. Professor Grizzle is a Fellow of the IEEE and IFAC. He received the Paper of the Year Award from the IEEE Vehicular Technology Society in 1993, the George S. Axelby Award in 2002, the Control Systems Technology Award in 2003, the Bode Prize in 2012, the IEEE Transactions on Control Systems Technology Outstanding Paper Award in 2014, and the  IEEE Transactions on Automation Science and Engineering, Googol Best New Application Paper Award in 2019. His work on bipedal locomotion has been the object of numerous plenary lectures and has been featured on CNN, ESPN, Discovery Channel, The Economist, Wired Magazine, Discover Magazine, Scientific American and Popular Mechanics.
\end{IEEEbiography}

\EOD

\end{document}